\documentclass[letterpaper]{article} 
\usepackage{aaai2026}  
\nocopyright
\usepackage{times}  
\usepackage{helvet}  
\usepackage{courier}  
\usepackage[hyphens]{url}  
\usepackage{graphicx} 
\usepackage{natbib}  
\usepackage{caption} 
\usepackage{algorithm}
\usepackage{algorithmic}

\usepackage{newfloat}
\usepackage{listings}
\DeclareCaptionStyle{ruled}{labelfont=normalfont,labelsep=colon,strut=off} 
\floatstyle{ruled}
\newfloat{listing}{tb}{lst}{}
\floatname{listing}{Listing}
\usepackage{amsmath,amsfonts}
\usepackage{amssymb}
\usepackage{threeparttable}
\usepackage{multirow}
\usepackage{booktabs}

\title{C3RL: Rethinking the Combination of Channel-independence and Channel-mixing from Representation Learning}
\author{
    Shusen Ma\textsuperscript{\rm 1}, Yun-Bo Zhao\textsuperscript{\rm 1}\thanks{Corresponding author.} and
    Yu Kang\textsuperscript{\rm 1}
}
\affiliations{
    \textsuperscript{\rm 1}University of Science and Technology of China\\

    mss0913@mail.ustc.edu.cn, ybzhao@ustc.edu.cn
}

\usepackage{bibentry}

\begin{document}

\maketitle

\begin{abstract}
Multivariate time series forecasting has drawn increasing attention due to its practical importance. Existing approaches typically adopt either channel-mixing (CM) or channel-independence (CI) strategies. CM strategy can capture inter-variable dependencies but fails to discern variable-specific temporal patterns. CI strategy improves this aspect but fails to fully exploit cross-variable dependencies like CM. Hybrid strategies based on feature fusion offer limited generalization and interpretability. To address these issues, we propose C3RL, a novel representation learning framework that jointly models both CM and CI strategies. Motivated by contrastive learning in computer vision, C3RL treats the inputs of the two strategies as transposed views and builds a siamese network architecture: one strategy serves as the backbone, while the other complements it. By jointly optimizing contrastive and prediction losses with adaptive weighting, C3RL balances representation and forecasting performance. Extensive experiments on seven models show that C3RL boosts the best-case performance rate to 81.4\% for models based on CI strategy and to 76.3\% for models based on CM strategy, demonstrating strong generalization and effectiveness.
\end{abstract}

 \begin{links}
     \link{Code}{https://github.com/SSMa913/NICLab-C3RL}
 \end{links}

\section{Introduction} \label{Introduction}

Multivariate time series forecasting (MTSF) has a wide range of real-world applications, including electricity consumption prediction \cite{qureshi2024deep}, traffic flow forecasting \cite{kong2024spatio}, and vital signal prediction \cite{10561458}. Multivariate time series (MTS) data are rich in features, capturing both the temporal dynamics of individual variables and the inter-variable correlations. These features encode not only evolving trends but also intricate entanglements among variables. Effectively extracting and modeling these representations is critical to improving forecasting accuracy and remains a central challenge in MTSF research.

Currently, two dominant input processing strategies are widely adopted: \textit{channel-mixing} (CM) and \textit{channel-independence} (CI) strategies. CM treats multivariate observations at each time step as a single token, focusing on temporal dependencies. In contrast, CI approaches treat each variable’s sequence as an independent token. Within this category, we identify two subtypes: \textit{explicit channel independence} (ECI), where variables are sequentially fed into a shared backbone to learn temporal patterns of diverse variables \cite{Yuqietal-2023-PatchTST}, and \textit{implicit channel independence} (ICI), where the sequence of shape (L, N) is transposed to (N, L) before being input as a whole \cite{ma2024multivariate, liu2024itransformer}.

Recent studies suggest that CI models often outperform CM ones in terms of prediction accuracy \cite{liu2024itransformer, zeng2023transformers}. However, relying solely on one strategy may fail to capture the full spectrum of temporal dependencies, especially for heterogeneous datasets \cite{ahamed2024timemachine}. As a result, recent methods attempt to combine both strategies \cite{ahamed2024timemachine, liang2024bi}, primarily through feature fusion \cite{fan2025dc}. Nonetheless, these fusion-based approaches often learn a single-task mapping focused on label prediction rather than robust representation learning. This leads to limited generalization, reduced interpretability, and deteriorated performance on unseen patterns.

Contrastive learning, as a form of self-supervised learning, aims to learn meaningful feature representations by modeling the similarity and dissimilarity between data samples. Extensive studies \cite{yue2022ts2vec, woo2022cost} have shown its effectiveness in learning useful temporal representations for time series forecasting. However, these studies predominantly rely on a single data processing strategy, especially the CM strategy. Therefore, integrating contrastive learning with both processing strategies is a promising direction. Such integration not only enhances the model's adaptability to diverse data types and scenarios but also improves its representational capacity. To achieve this, several key challenges must be addressed:

\textit{1) How to Reduce Training Cost while Avoiding Collapsing Solutions?} 
A common issue in contrastive learning is representation collapse, where the model maps all inputs to identical or near-identical embeddings due to insufficient constraints on representational diversity \cite{chen2021exploring}. SimCLR \cite{chen2020simple} mitigates this via negative sample pairs, but at the cost of increased computation. BYOL \cite{grill2020bootstrap} uses only positive pairs, yet avoids collapse due to the use of a momentum encoder.

\textit{2) How to Construct Meaningful Positive Sample Pairs?} 
In computer vision, positive sample pairs are typically created by applying data augmentations (e.g., rotation, cropping) to the same image \cite{he2020momentum}, while samples from different images are treated as negatives \cite{chen2020simple}. However, in the context of time series forecasting, how to effectively construct positive pairs remains underexplored, especially when jointly considering CM and CI strategies. There is a lack of systematic investigation into leveraging these two complementary views to define semantically meaningful sample pairs for contrastive learning.

\textit{3) How to Design an Effective Loss Function?} 
Contrastive loss functions guide the model to pull together positive pairs and push apart negative ones, thereby learning discriminative features. However, since our ultimate goal is forecasting, the prediction loss—which measures the discrepancy between predicted and ground truth values—must also be incorporated. Balancing representation learning (via contrastive loss) and prediction accuracy (via forecasting loss) is thus a central challenge in loss design.

\begin{figure}[t]
	\centering
	\includegraphics[width=0.48\textwidth]{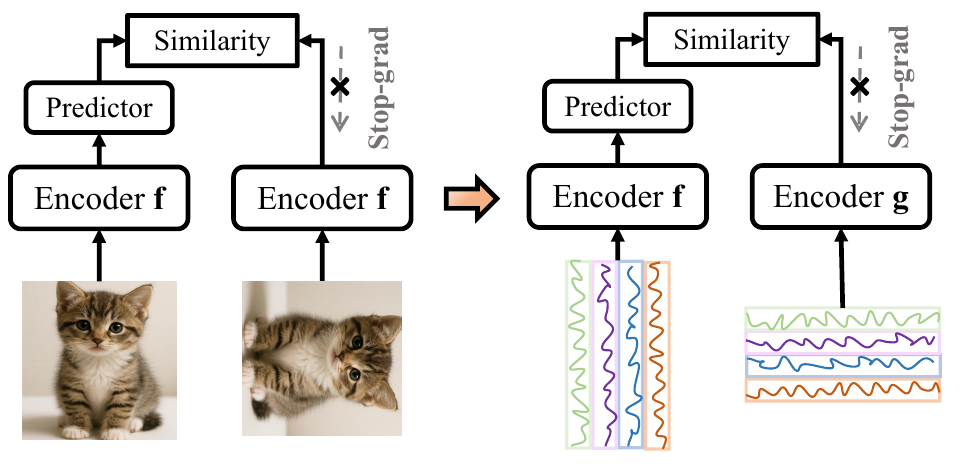}
	\caption{The application of the simple siamese networks (SimSiam) from the images (left) to the time series (right).}
	\label{simsiam}
\end{figure}

To address the aforementioned limitations, we rethink the \textbf{C}ombination of \textbf{C}hannel-independence and \textbf{C}hannel-mixing from a \textbf{R}epresentation \textbf{L}earning perspective, and propose a unified framework, \textbf{C3RL}. 

\textit{1) SimSiam-inspired Architecture Design.} 
To avoid collapsing solutions without increasing training cost by adding more data, C3RL adopts the SimSiam architecture \cite{chen2021exploring} as its design inspiration. As shown in the left part of Figure \ref{simsiam}, SimSiam effectively avoids representational collapse by applying a stop-gradient on one branch, eliminating the need for negative samples. In SimSiam, both inputs share the same encoder. However, in C3RL, since the time series inputs differ in dimensional structure due to transposition, the same encoder cannot process both views consistently. To solve this, we introduce a siamese encoder $g$ corresponding to the backbone encoder $f$, as illustrated on the right side of Figure \ref{simsiam}. The siamese encoder is designed to preserve the architecture of $f$, with internal feature dimensions adjusted only to accommodate the input shape. 

\textit{2) Channel Views as Positive Sample Pairs.}
Upon analysis, we observe that the inputs under the CM and ICI strategies can be viewed as transposed versions of each other. This rotation-like transformation aligns with the semantics of positive pairs in SimSiam, and thus, they can be naturally treated as positive examples within our contrastive learning setup (see Figure~\ref{simsiam}, right). The application of C3RL to the ECI can see the \textbf{Appendix}.

\textit{3) Joint Training for Representation and Forecasting.}
To enhance representation learning while maintaining strong performance on downstream forecasting tasks, C3RL adopts a joint optimization strategy, where SimSiam-based contrastive learning and supervised prediction are trained simultaneously. Given that the importance of contrastive and predictive signals can vary across datasets and tasks, we introduce adaptive weights to balance the contrastive loss and the prediction loss dynamically. This enables the model to adaptively trade off between representational richness and task-specific accuracy, leading to superior generalization across diverse scenarios. 

The main contributions of this paper are summarized as follows:
\begin{itemize}
\item We propose a novel paradigm, C3RL, based on representation learning, which unifies CM and CI strategies within a new framework. This approach significantly improves the performance and enhances the representation ability of most mainstream forecasting models and offers a new modeling perspective for future research.
\item We design a new loss function that integrates contrastive loss and forecasting loss, enabling dynamic adjustment of their relative weights. This allows the model to maintain an optimal balance between representation quality and forecasting accuracy across different tasks.
\item We validate the effectiveness of C3RL on nine publicly available real-world datasets using five CI models and two CM models. Specifically, C3RL boosts the best-case performance rate from 43.6\% to 81.4\% for CI models, and from 23.8\% to 76.3\% for CM models, demonstrating strong generalization and effectiveness.
\end{itemize}

\begin{figure*}[t]
	\centering
	\includegraphics[width=0.95\textwidth]{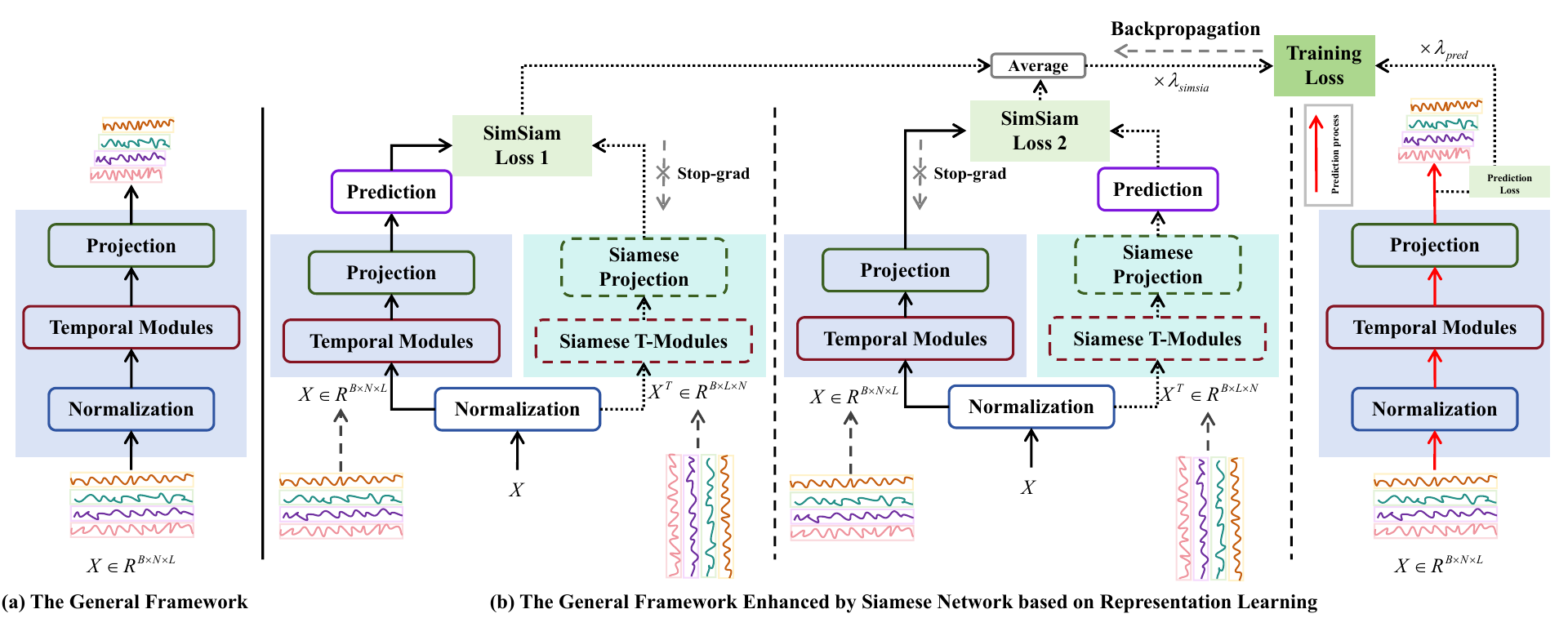}
	\caption{The pipeline of the combination of CI and CM by representation learning (C3RL). The T-Modules denotes the temporal modules}
	\label{pipeline}
\end{figure*}

\section{Related Work}
Traditional time series forecasting models, such as ARIMA \cite{zhang2003time} and SVM \cite{cao2003support}, are simple and effective in certain scenarios. However, they often struggle to capture the complex nonlinear dependencies among multiple variables. With the increasing availability of data and the rapid development of deep learning techniques, various neural network architectures, such as Recurrent Neural Networks \cite{hewamalage2021recurrent} and Convolutional Neural Networks \cite{ma2023tcln}, have been increasingly applied to time series forecasting tasks. In particular, the emergence of the Transformer \cite{NIPS2017_3f5ee243} has sparked extensive research efforts aimed at leveraging its strength in modeling long-term dependencies to improve forecasting performance.
Informer \cite{zhou2021informer} stands out as a milestone in MTSF, adopting a typical CM strategy and introducing a generative-style decoder that enables long-sequence prediction in a single forward pass. Building on this paradigm, numerous models have been proposed, including Autoformer \cite{wu2021autoformer}, FEDformer \cite{zhou2022fedformer}, and Pyraformer \cite{liu2022pyraformer}. However, the CM strategy, which treats the multivariate information at each time step as a single token, often leads to excessive coupling between variables and undermines predictive accuracy.
To address this issue, PCDformer \cite{ma2024multivariate} introduces a parallel convolutional mechanism to independently model each variable, reducing inter-variable interference. iTransformer \cite{liu2024itransformer} employs an inversion of input data and relies solely on the Transformer encoder, achieving competitive performance. In parallel, lightweight models based on multi-layer perceptrons (MLPs), such as DLinear \cite{zeng2023transformers} and RLinear \cite{li2023revisiting}, have demonstrated the ability to outperform some Transformer-based approaches. More recently, the advent of Mamba, with its State Space Model (SSM) structure and parallel scan mechanism, has shown promising results in both computational efficiency and forecasting accuracy \cite{ma2024fmamba}.
While these methods improve upon the traditional CM strategy by better highlighting variable-specific features, they still differ from the strategy adopted in PatchTST \cite{Yuqietal-2023-PatchTST}, where each variable is modeled in a completely independent manner, referred to as the ECI strategy. In contrast, we term the aforementioned approaches as adopting an ICI strategy.

Although CI modeling generally outperforms CM-based models, relying solely on a single strategy is insufficient to fully capture the complex patterns embedded in MTS data \cite{ahamed2024timemachine}. As a result, recent studies have explored hybrid approaches that combine CM and CI strategies to build more generalizable forecasting frameworks. For example, TimeMachine \cite{ahamed2024timemachine} utilizes pure SSM modules to extract potential cross-channel correlations; Bi-Mamba+ \cite{liang2024bi} introduces a Series-Relation-Aware module to dynamically select between CI and CM strategies; and DC-Mamba \cite{fan2025dc} integrates Mamba with linear attention mechanisms to fuse features extracted by different strategies.
However, most of these works focus primarily on improving predictive accuracy through strategy fusion, often overlooking the model's representation capacity and interpretability. Therefore, it remains a promising direction to investigate how to effectively integrate CM and CI strategies into a unified framework that not only enhances forecasting performance but also improves the model's representational power and interpretability.

\section{Methodology}

In this work, we rethink and propose the C3RL, a novel paradigm for model enhancement that can be seamlessly applied to mainstream MTSF models. The overall pipeline of C3RL is illustrated in Figure~\ref{pipeline}.
As shown in Figure~\ref{pipeline}(a), a general time series forecasting model typically consists of a normalization layer (e.g., Layer normalization \cite{ba2016layer} or RevIN \cite{kim2021reversible}), temporal modules (such as linear-based, Mamba-based, or Transformer-based architectures), and a projection layer (i.e., prediction head).
As introduced in Introduction, we extend this baseline framework by designing a siamese-network architecture, where each branch adopts a different channel strategy. Furthermore, we introduce a joint loss function based on weight to balance the model’s representational and predictive capabilities. The detailed pipeline of our proposed framework is shown in Figure~\ref{pipeline}(b).

\subsection{Preliminary}
Let $X = [x_1, x_2, \ldots, x_L] \in \mathbb{R}^{L \times N}$ denote an MTS data, where $L$ is the total number of time steps and $N$ represents the number of variables (or channels). Each observation at time step $t$ is denoted by $x_t \in \mathbb{R}^N$. The MTSF task aims to learn a mapping function $f$ that predicts the future sequence $Y = [\hat{x}_{L+1}, \ldots, \hat{x}_{L+P}] \in \mathbb{R}^{P \times N}$, where $P$ is the forecasting horizon. 
Specifically, models that adopt the CM strategy take inputs in the form of $X = [x_1, x_2, \ldots, x_L] \in \mathbb{R}^{L \times N}$. In contrast, models that use an ICI strategy receive inputs as $X = [x_1, x_2, \ldots, x_N] \in \mathbb{R}^{N \times L}$. For models with an ECI strategy, the multivariate input needs to be completely separated beforehand, transforming the input from $X \in \mathbb{R}^{B \times N \times L}$ to $X \in \mathbb{R}^{(B \cdot N) \times 1 \times L}$, where $B$ is the batch size.

\subsection{Siamese Network} 
\label{Siamese Network}

The core idea of C3RL is to enhance the representational and predictive capabilities of existing backbone models by constructing a corresponding Siamese network, as shown in Figure~\ref{pipeline}(b). This Siamese network comprises two components: Siamese Temporal Modules (T-Modules) and a Siamese Projection. The T-Modules are designed by replicating the temporal structure of the backbone while adjusting only the internal feature dimensions to meet input requirements, avoiding the need for newly engineered feature extractors. The Siamese Projection is implemented with a minimal MLP, aiming to align its output features with those of the backbone's projection head using the fewest possible layers.
\begin{figure}[t]
	\centering
	\includegraphics[width=0.46\textwidth]{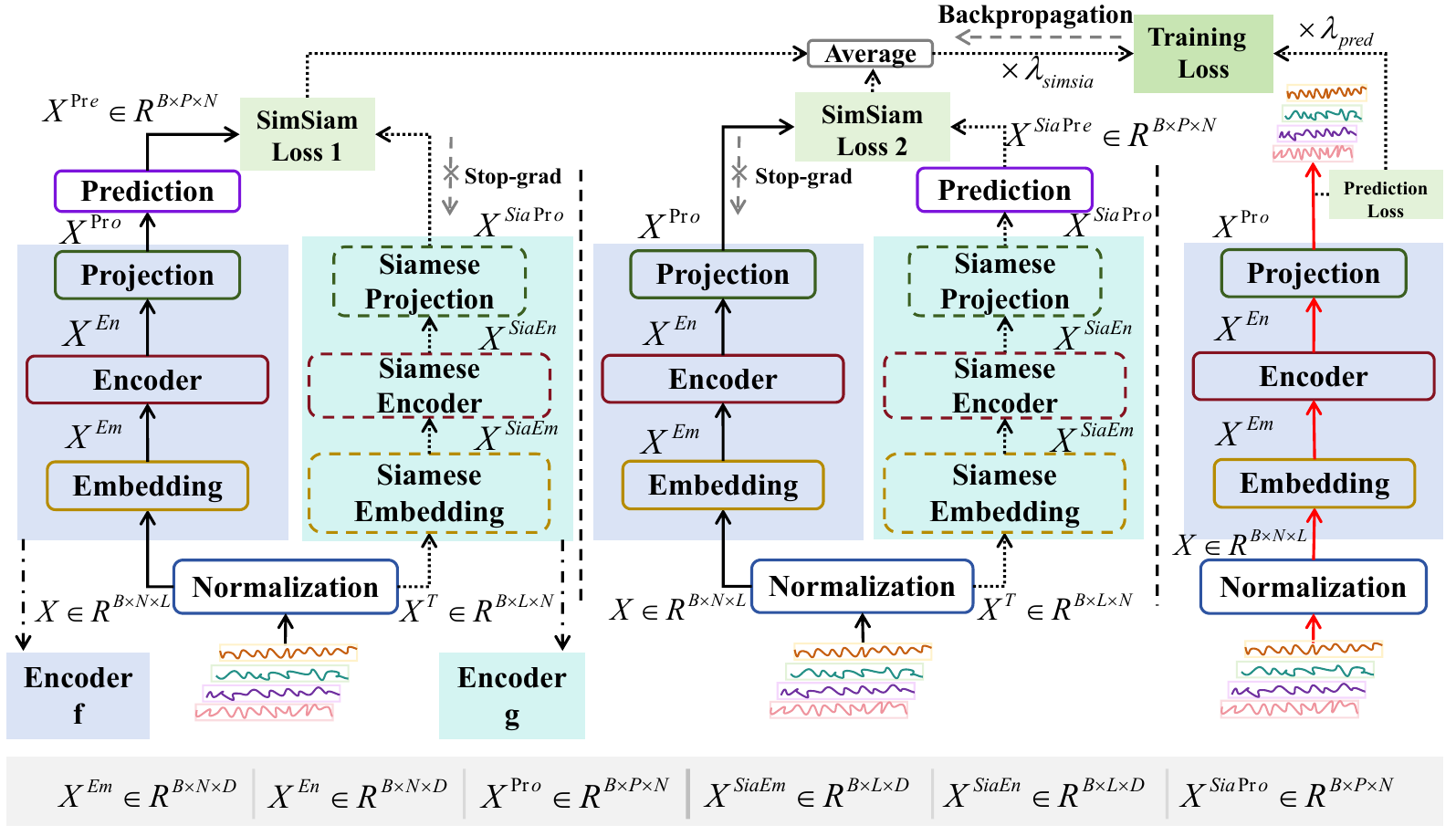}
	\caption{The application of C3RL to the iTransformer.}
	\label{iTransformer-contrast-learning}
\end{figure}
For clarity, Figure~\ref{iTransformer-contrast-learning} presents the construction of a Siamese network by incorporating C3RL into iTransformer (more examples shown in the \textbf{Appendix}):

\textit{Step 1:} In iTransformer, the Embedding and Encoder layers are considered as part of the T-Modules. These components can be directly transferred to construct the Siamese Embedding and Siamese Encoder layers.
\textit{Step 2:} Since the input dimensions on the Siamese side differ, the internal parameters of the Siamese Embedding layer must be adjusted accordingly to ensure proper data flow. Specifically, the input feature dimension of the original Embedding layer (implemented via a linear layer) is $L$, so the linear layer is defined as $nn.Linear(D_{in}, D)$ with $D_{in} = L$. In contrast, the input feature dimension for the Siamese Embedding layer is $N$, and thus $D_{in}$ should be set to $N$ in the corresponding linear layer.
\textit{Step 3:} The output of the Siamese Embedding layer, which serves as the input to the Siamese Encoder, has the same last-dimensional size $D$ as the input to the original Encoder. Therefore, the internal structure of the Siamese Encoder remains unchanged.
\textit{Step 4:} The output of the original Projection layer is denoted as $X^{Pro} \in \mathbb{R}^{B \times P \times N}$. To align the output of the Siamese Encoder, $X^{SiaEn} \in \mathbb{R}^{B \times L \times D}$, with $X^{Pro}$ in feature space, we introduce a Siamese Projection layer, as illustrated in Figure~\ref{siamese-projection}(a).

\subsection{Training and Prediction}

\begin{figure}[t]
	\centering
	\includegraphics[width=0.38\textwidth]{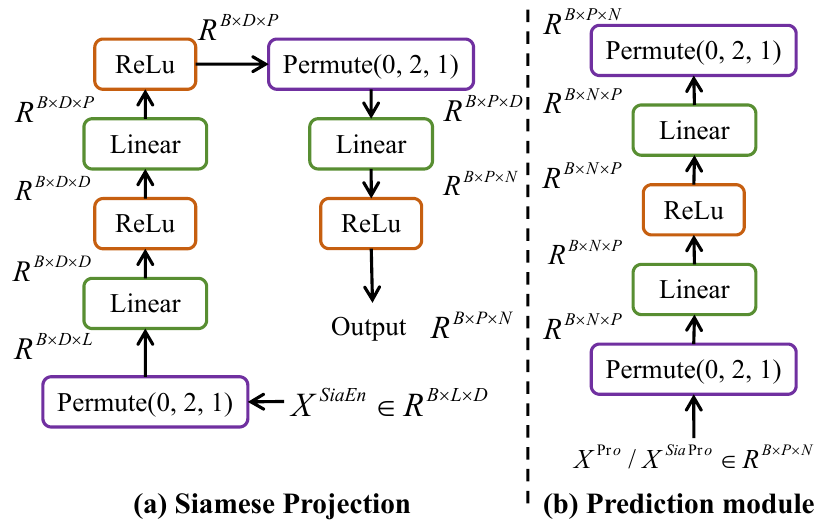}
	\caption{The structure of the Siamese Projection (left) and Prediction module (right).}
	\label{siamese-projection}
\end{figure}

In this section, we introduce the formulation of the training loss and the prediction process of the model. As illustrated in Figure~\ref{iTransformer-contrast-learning}, $X$ and $X^T$ are used as the inputs to the Encoder $f$ and $g$, respectively. We define the output of the Encoder $f$ and Encoder $g$ as $X^{Pro} \triangleq Enc_f(X)$ and $X^{SiaPro} \triangleq Enc_g(X^T)$, respectively. Prediction module is used to transform the output of one of the Encoder and match it to the view of the other Encoder. The structure of the Prediction module is shown as Figure~\ref{siamese-projection}(b).
As shown on the left side of Figure~\ref{iTransformer-contrast-learning}, we define $X^{Pre} \triangleq P(X^{Pro})$, where $P(\cdot)$ denotes the Prediction module.
Following \cite{chen2021exploring}, we minimize the negative cosine similarity of the $X^{Pre}$ and $X^{SiaPro}$:
\begin{equation}
    \mathcal{D}(X^{Pre}, X^{SiaPro}) = -\frac{X^{Pre}}{\|X^{Pre}\|_2} \cdot \frac{X^{SiaPro}}{\|X^{SiaPro}\|_2}, \label{neg_cos_sim}
\end{equation}
where $\| \cdot \|_2$ denotes the $l_2$-norm.
Following \cite{grill2020bootstrap}, we define a symmetrized loss as:
\begin{equation}
    \mathcal{L}_{simsia} = \frac{1}{2} \cdot \mathcal{D}(X^{Pre}, X^{SiaPro}) + \frac{1}{2} \cdot \mathcal{D}(X^{Pro}, X^{SiaPre}), \label{loss_simsia}
\end{equation}
where $X^{SiaPre}$ represents the matching of the Encoder $g$ to the Encoder $f$ by the Prediction module shown in the middle of Figure~\ref{iTransformer-contrast-learning}. Following \cite{chen2021exploring}, we adopt the stop-gradient (stop-grad) operation to avoid the collapsing solutions. Therefore, we need to modify the equation (\ref{neg_cos_sim}) to:
\begin{equation}
\begin{split}
   & \mathcal{D}( X^{Pre}, \text{stop-grad} (X^{SiaPro}) ) \\
   & = -\frac{X^{Pre}}{\|X^{Pre}\|_2} \cdot \text{stop-grad}( \frac{X^{SiaPro}}{\|X^{SiaPro}\|_2} ), 
\end{split}
\end{equation}
which means that $X^{SiaPro}$ is excluded from gradient computation. Similarly, we redefine the equation (\ref{loss_simsia}) as:
\begin{equation}
\begin{split}
    & \mathcal{L}_{simsia} = \frac{1}{2} \cdot \mathcal{D}(X^{Pre}, \text{stop-grad} ( X^{SiaPro} ) ) \\
    & + \frac{1}{2} \cdot \mathcal{D}( \text{stop-grad} ( X^{Pro} ), X^{SiaPre}).
\end{split}
\end{equation}

\begin{table*}[ht]
\centering
\setlength{\tabcolsep}{1.3mm}
\scriptsize 
\begin{tabular}{l|r|cccc|cccc|cccc|cccc|cccc}
\toprule
\multicolumn{2}{c|}{Model} & \multicolumn{2}{c}{S-Mamba} & \multicolumn{2}{c|}{\textbf{+ C3RL}} & \multicolumn{2}{c}{DLinear} & \multicolumn{2}{c|}{\textbf{+ C3RL}} & \multicolumn{2}{c}{PatchTST} & \multicolumn{2}{c|}{\textbf{+ C3RL}} & \multicolumn{2}{c}{iTransformer} & \multicolumn{2}{c|}{\textbf{+ C3RL}} & \multicolumn{2}{c}{{RLinear}} & \multicolumn{2}{c}{\textbf{+ C3RL}} \\

\multicolumn{2}{c|}{Metric} & \multicolumn{1}{l}{MSE} &  {MAE} &  {MSE} &  {MAE} & \multicolumn{1}{l}{MSE} &  {MAE} &  {MSE} &  {MAE} & \multicolumn{1}{l}{MSE} &  {MAE} &  {MSE} &  {MAE} & \multicolumn{1}{l}{MSE} &  {MAE} & MSE & MAE & MSE & MAE & MSE & MAE\\

\midrule
\multirow{4}{*}{\rotatebox[origin=c]{90}{\textbf{ETTh1}}} & 96 & 0.388 & 0.406 & \textbf{0.386} & \textbf{0.405} & 0.384 & 0.405 & \textbf{0.374} & \textbf{0.395} & \textbf{0.375} & \textbf{0.399} & 0.376 & 0.400 & \textbf{0.387} & \textbf{0.405} & \textbf{0.387} & \textbf{0.405} & \textbf{0.366} & \textbf{0.391} & 0.367 & \textbf{0.391} \\

 & 192 & 0.445 & \textbf{0.441} & \textbf{0.442} & 0.442 & 0.443 & 0.450 & \textbf{0.408} & \textbf{0.418} & \textbf{0.414} & \textbf{0.421} & \textbf{0.414} & \textbf{0.421} & \textbf{0.441} & \textbf{0.436} & \textbf{0.441} & 0.438 & \textbf{0.403} & \textbf{0.412} & 0.404 & \textbf{0.412} \\
 
 & 336 & 0.490 & \textbf{0.465} & \textbf{0.485} & \textbf{0.465} & 0.447 & 0.448 & \textbf{0.443} & \textbf{0.444} & 0.431 & 0.435 & \textbf{0.429} & \textbf{0.434} & 0.491 & 0.462 & \textbf{0.485} & \textbf{0.461} & 0.420 & \textbf{0.423} & \textbf{0.418} & \textbf{0.423} \\
 
 & 720 & 0.506 & 0.497 & \textbf{0.501} & \textbf{0.494} & 0.504 & 0.515 & \textbf{0.472} & \textbf{0.489} & 0.450 & 0.466 & \textbf{0.445} & \textbf{0.463} & 0.509 & 0.494 & \textbf{0.503} & \textbf{0.491} & 0.442 & 0.456 & \textbf{0.431} & \textbf{0.449} \\
 
 \midrule
\multirow{4}{*}{\rotatebox[origin=c]{90}{\textbf{ETTh2}}} & 96 & 0.297 & 0.349 & \textbf{0.295} & \textbf{0.347} & 0.290 & 0.353 & \textbf{0.280} & \textbf{0.349} & \textbf{0.274} & 0.336 & \textbf{0.274} & \textbf{0.335} & 0.301 & 0.350 & \textbf{0.297} & \textbf{0.349} & 0.262 & 0.331 & \textbf{0.260} & \textbf{0.328} \\

 & 192 & 0.378 & 0.399 & \textbf{0.376} & \textbf{0.398} & 0.388 & 0.422 & \textbf{0.362} & \textbf{0.403} & \textbf{0.338} & \textbf{0.378} & \textbf{0.338} & 0.379 & 0.380 & 0.399 & \textbf{0.378} & \textbf{0.398} & 0.320 & 0.374 & \textbf{0.317} & \textbf{0.369} \\
 
 & 336 & 0.425 & 0.435 & \textbf{0.423} & \textbf{0.434} & 0.463 & 0.473 & \textbf{0.436} & \textbf{0.455} & 0.331 & 0.380 & \textbf{0.327} & \textbf{0.378} & 0.424 & \textbf{0.432} & \textbf{0.422} & \textbf{0.432} & 0.326 & 0.388 & \textbf{0.324} & \textbf{0.385} \\
 
 & 720 & 0.432 & 0.448 & \textbf{0.431} & \textbf{0.447} & 0.733 & 0.606 & \textbf{0.667} & \textbf{0.580} & 0.379 & 0.421 & \textbf{0.375} & \textbf{0.418} & 0.430 & \textbf{0.447} & \textbf{0.429} & \textbf{0.447} & 0.425 & 0.449 & \textbf{0.412} & \textbf{0.443} \\
 \midrule
 
\multirow{4}{*}{\rotatebox[origin=c]{90}{\textbf{ETTm1}}} & 96 & \textbf{0.331} & \textbf{0.368} & 0.332 & 0.369 & 0.301 & 0.345 & \textbf{0.300} & \textbf{0.344} & 0.292 & 0.343 & \textbf{0.289} & \textbf{0.342} & \textbf{0.342} & 0.377 & \textbf{0.342} & \textbf{0.376} & \textbf{0.301} & \textbf{0.343} & 0.302 & \textbf{0.343} \\

 & 192 & 0.378 & \textbf{0.393} & \textbf{0.377} & \textbf{0.393} & \textbf{0.336} & \textbf{0.366} & 0.337 & 0.367 & \textbf{0.331} & \textbf{0.369} & 0.332 & \textbf{0.369} & 0.383 & 0.396 & \textbf{0.381} & \textbf{0.394} & 0.341 & 0.367 & \textbf{0.338} & \textbf{0.365} \\
 
 & 336 & 0.410 & \textbf{0.414} & \textbf{0.408} & \textbf{0.414} & \textbf{0.372} & \textbf{0.389} & 0.375 & 0.391 & \textbf{0.365} & 0.392 & 0.366 & \textbf{0.391} & 0.418 & 0.418 & \textbf{0.416} & \textbf{0.417} & 0.374 & 0.386 & \textbf{0.371} & \textbf{0.384} \\
 
 & 720 & \textbf{0.474} & \textbf{0.451} & 0.475 & 0.453 & \textbf{0.427} & \textbf{0.423} & 0.432 & 0.427 & 0.421 & 0.425 & \textbf{0.420} & \textbf{0.424} & 0.487 & 0.457 & \textbf{0.484} & \textbf{0.454} & 0.430 & 0.418 & \textbf{0.426} & \textbf{0.415} \\
 \midrule
 
\multirow{4}{*}{\rotatebox[origin=c]{90}{\textbf{ETTm2}}} & 96 & 0.182 & 0.266 & \textbf{0.179} & \textbf{0.264} & 0.172 & \textbf{0.267} & \textbf{0.171} & 0.268 & 0.164 & 0.254 & \textbf{0.163} & \textbf{0.252} & 0.186 & 0.272 & \textbf{0.183} & \textbf{0.267} & \textbf{0.164} & 0.253 & \textbf{0.164} & \textbf{0.252} \\

 & 192 & 0.252 & 0.313 & \textbf{0.248} & \textbf{0.309} & 0.237 & \textbf{0.314} & \textbf{0.235} & 0.317 & 0.221 & \textbf{0.292} & \textbf{0.220} & \textbf{0.292} & 0.254 & 0.314 & \textbf{0.251} & \textbf{0.312} & \textbf{0.219} & \textbf{0.290} & \textbf{0.219} & \textbf{0.290} \\
 
 & 336 & 0.313 & 0.349 & \textbf{0.311} & \textbf{0.348} & \textbf{0.295} & \textbf{0.359} & 0.313 & 0.373 & 0.278 & \textbf{0.329} & \textbf{0.277} & 0.330 & 0.316 & \textbf{0.351} & \textbf{0.315} & \textbf{0.351} & \textbf{0.273} & \textbf{0.326} & 0.274 & \textbf{0.326} \\
 
 & 720 & \textbf{0.413} & \textbf{0.405} & \textbf{0.413} & \textbf{0.405} & 0.427 & 0.439 & \textbf{0.392} & \textbf{0.411} & 0.367 & 0.385 & \textbf{0.362} & \textbf{0.382} & 0.414 & 0.407 & \textbf{0.410} & \textbf{0.405} & \textbf{0.366} & 0.385 & 0.367 & \textbf{0.384} \\
 \midrule
 
\multirow{4}{*}{\rotatebox[origin=c]{90}{\textbf{Exchange}}} & 96 & \textbf{0.086} & \textbf{0.206} & 0.088 & 0.208 & 0.085 & 0.209 & \textbf{0.082} & \textbf{0.202} & \textbf{0.093} & \textbf{0.213} & \textbf{0.093} & \textbf{0.213} & \textbf{0.086} & \textbf{0.206} & 0.087 & 0.208 & \textbf{0.090} & \textbf{0.209} & \textbf{0.090} & \textbf{0.209} \\

 & 192 & 0.182 & 0.304 & \textbf{0.180} & \textbf{0.302} & 0.162 & 0.296 & \textbf{0.160} & \textbf{0.291} & \textbf{0.194} & \textbf{0.314} & \textbf{0.194} & \textbf{0.314} & 0.181 & 0.304 & \textbf{0.178} & \textbf{0.302} & 0.193 & 0.311 & \textbf{0.186} & \textbf{0.305} \\
 
 & 336 & \textbf{0.331} & \textbf{0.417} & \textbf{0.331} & \textbf{0.417} & 0.333 & 0.441 & \textbf{0.319} & \textbf{0.435} & 0.355 & 0.436 & \textbf{0.352} & \textbf{0.433} & \textbf{0.338} & \textbf{0.422} & \textbf{0.338} & \textbf{0.422} & 0.362 & 0.435 & \textbf{0.361} & \textbf{0.434} \\
 
 & 720 & \textbf{0.858} & \textbf{0.699} & 0.862 & 0.701 & 0.898 & 0.725 & \textbf{0.888} & \textbf{0.721} & \textbf{0.903} & \textbf{0.712} & 0.904 & 0.713 & 0.853 & \textbf{0.696} & \textbf{0.852} & 0.697 & 0.923 & \textbf{0.719} & \textbf{0.919} & \textbf{0.719} \\
 \midrule

\multirow{4}{*}{\rotatebox[origin=c]{90}{\textbf{Illness}}} & 24 & 1.849 & 0.865 & \textbf{1.808} & \textbf{0.854} & \textbf{2.280} & \textbf{1.061} & 2.289 & 1.064 & \textbf{1.401} & \textbf{0.738} & 1.522 & 0.815 & 2.358 & 1.063 & \textbf{2.345} & \textbf{1.051} & 2.306 & 1.037 & \textbf{2.303} & \textbf{1.032} \\

 & 36 & 2.188 & 1.006 & \textbf{1.776} & \textbf{0.877} & \textbf{2.235} & \textbf{1.059} & 2.274 & 1.063 & 1.452 & 0.839 & \textbf{1.317} & \textbf{0.781} & 2.184 & 1.002 & \textbf{2.182} & \textbf{0.999} & 2.173 & 1.014 & \textbf{2.069} & \textbf{0.977} \\
 
 & 48 & \textbf{1.708} & \textbf{0.875} & 1.787 & 0.894 & 2.298 & \textbf{1.079} & \textbf{2.285} & 1.087 & 1.679 & 0.859 & \textbf{1.498} & \textbf{0.817} & \textbf{2.124} & 1.019 & 2.161 & \textbf{1.018} & 2.184 & 1.005 & \textbf{2.151} & \textbf{0.996} \\
 
 & 60 & 2.338 & 1.005 & \textbf{2.068} & \textbf{0.953} & 2.573 & 1.157 & \textbf{2.420} & \textbf{1.111} & 1.621 & 0.888 & \textbf{1.395} & \textbf{0.798} & 2.165 & 1.032 & \textbf{2.061} & \textbf{0.991} & 2.087 & 0.983 & \textbf{2.080} & \textbf{0.981} \\

 \bottomrule
\end{tabular}
\caption{Results on base models based on CI. The better performance in each setting is shown in \textbf{bold}.
}
\label{channel_independence_c3rl}
\end{table*}

$\mathcal{L}_{simsia}$ is a contrastive loss function designed to pull together features extracted using different data augmentation strategies (ICI and ECI) in the feature space. This encourages the model to learn richer temporal patterns and enhances its representational capacity. However, since the ultimate objective is forecasting, the construction of the training loss must also incorporate a prediction error term. To balance the influence of contrastive and error-based losses, we assign them different weights and dynamically adjust their contributions in the training. This strategy aims to enhance the model's representation ability while improving the model’s final predictive performance. The overall training loss is defined as follows:
\begin{equation}
    \mathcal{L}_{\rm total} = \lambda_{simsia} \cdot \mathcal{L}_{simsia} + \lambda_{pred} \cdot \mathcal{L}_{pred},
\end{equation}
where $\lambda_{simsia}$ and $\lambda_{pred}$ are tunable hyperparameters. The inference (prediction) process of the trained model, indicated by the red arrow on the right side of Figure~\ref{iTransformer-contrast-learning}, remains consistent with that of iTransformer.

\section{Experiments}

\subsection{Experimental Settings}

\textbf{Datasets.}
We conduct extensive experiments on nine publicly available and widely used datasets: ETT (ETTh1, ETTh2, ETTm1 and ETTm2) \cite{zhou2021informer}, Exchange \cite{lai2018modeling}, Weather, Electricity, Traffic, and Illness \cite{wu2021autoformer}. 

\textbf{Evaluation metrics.}
We adopt two widely used evaluation metrics from existing benchmarking protocols: Mean Squared Error (MSE) and Mean Absolute Error (MAE), to compare the performance of mainstream models with their enhanced versions integrated with C3RL.

\textbf{Compared baselines.}
Since models based on the CI strategy have achieved state-of-the-art performance, most recent studies have been built upon this approach \cite{ma2024fmamba, liang2024bi, fan2025dc}. Therefore, our main comparisons and analyses focus on models following this strategy. Specifically, we select three representative models constructed under different neural network paradigms: MLP-based (DLinear \cite{zeng2023transformers} and RLinear \cite{li2023revisiting}), Transformer-based (iTransformer \cite{liu2024itransformer} and PatchTST \cite{Yuqietal-2023-PatchTST}), and Mamba-based (S-Mamba \cite{wang2025mamba}). In addition, to evaluate the generalization capability of C3RL, we include two representative models based on the CM strategy: Informer \cite{zhou2021informer} and Autoformer \cite{wu2021autoformer}.

\textbf{Experimental details.}
All experiments are implemented in PyTorch and conducted on a single NVIDIA GeForce RTX 3090 GPU. To ensure fair comparisons, we reimplement all baselines and their C3RL-enhanced variants under the same hardware and software settings. The hyperparameters for baselines follow their official implementations. For the baseline implementations, we kept the original random seed values unchanged. For those without a predefined seed, we uniformly set the random seed to 2025.

\subsection{Results and Analysis}

\begin{table}[t]
\centering
\setlength{\tabcolsep}{1.3mm}
\scriptsize 
\begin{tabular}{l|r|cccc|cccc}
\toprule
\multicolumn{2}{c|}{Model} & \multicolumn{2}{c}{Informer} & \multicolumn{2}{c|}{\textbf{+ C3RL}} & \multicolumn{2}{c}{Autoformer} & \multicolumn{2}{c}{\textbf{+ C3RL}}  \\

\multicolumn{2}{c|}{Metric} & \multicolumn{1}{l}{MSE} &  {MAE} &  {MSE} &  {MAE} & \multicolumn{1}{l}{MSE} &  {MAE} &  {MSE} &  {MAE} \\

\midrule
\multirow{4}{*}{\rotatebox[origin=c]{90}{\textbf{ETTh1}}} & 24(24) & 0.567 & 0.550 & \textbf{0.486} & \textbf{0.500} & 0.381 & \textbf{0.422} & \textbf{0.376} & 0.424  \\

 & 48(48) & 0.674 & \textbf{0.613} & \textbf{0.657} & 0.617 & 0.391 & 0.419 & \textbf{0.381} & \textbf{0.413}  \\
 
 & 168(168) & 1.049 & 0.831 & \textbf{1.000} & \textbf{0.777} & 0.502 & 0.483 & \textbf{0.458} & \textbf{0.457}  \\
 
 & 336(336) & 1.434 & 0.990 & \textbf{1.152} & \textbf{0.865} & 0.511 & 0.488 & \textbf{0.492} & \textbf{0.480}  \\

 & 720(720) & 1.388 & 0.942 & \textbf{1.450} & \textbf{0.940} & \textbf{0.499} & \textbf{0.501} & 0.500 & 0.504  \\
 \midrule
 
\multirow{4}{*}{\rotatebox[origin=c]{90}{\textbf{ETTh2}}} & 24(24) & 0.761 & 0.669 & \textbf{0.408} & \textbf{0.503} & \textbf{0.259} & \textbf{0.339} & 0.278 & 0.360  \\

 & 48(48) & \textbf{1.863} & \textbf{1.096} & 1.923 & 1.097 & 0.314 & \textbf{0.374} & \textbf{0.312} & 0.376  \\
 
 & 168(168) & 4.152 & 1.727 & \textbf{2.726} & \textbf{1.391} & 0.491 & 0.471 & \textbf{0.427} & \textbf{0.432}  \\
 
 & 336(336) & 3.736 & 1.605 & \textbf{2.942} & \textbf{1.433} & 0.483 & 0.486 & \textbf{0.459} & \textbf{0.468}  \\

 & 720(720) & 3.821 & 1.648 & \textbf{3.070} & \textbf{1.509} & \textbf{0.478} & \textbf{0.487} & 0.483 & 0.500  \\
 \midrule
 
\multirow{4}{*}{\rotatebox[origin=c]{90}{\textbf{ETTm1}}} & 24(24) & 0.408 & 0.421 & \textbf{0.325} & \textbf{0.376} & \textbf{0.396} & \textbf{0.425} & 0.409 & 0.426  \\

 & 48(48) & 0.444 & 0.469 & \textbf{0.416} & \textbf{0.421} & 0.476 & 0.455 & \textbf{0.467} & \textbf{0.454}  \\
 
 & 96(96) & 0.520 & 0.528 & \textbf{0.508} & \textbf{0.509} & 0.467 & \textbf{0.452} & \textbf{0.443} & 0.455  \\
 
 & 288(288) & 0.907 & 0.746 & \textbf{0.838} & \textbf{0.689} & 0.584 & 0.519 & \textbf{0.574} & \textbf{0.509}  \\

 & 672(672) & 1.030 & 0.822 & \textbf{0.991} & \textbf{0.791} & 0.552 & \textbf{0.508} & \textbf{0.540} & 0.510  \\
 \midrule
 
\multirow{4}{*}{\rotatebox[origin=c]{90}{\textbf{Weather}}} & 24(48) & 0.108 & 0.179 & \textbf{0.107} & \textbf{0.173} & 0.230 & \textbf{0.311} & \textbf{0.230} & 0.316  \\

 & 48(96) & 0.216 & 0.314 & \textbf{0.191} & \textbf{0.286} & 0.260 & 0.331 & \textbf{0.246} & \textbf{0.318}  \\
 
 & 168(192) & 0.330 & 0.374 & \textbf{0.294} & \textbf{0.356} & 0.344 & 0.393 & \textbf{0.295} & \textbf{0.354}  \\
 
 & 336(336) & 0.443 & \textbf{0.453} & \textbf{0.414} & 0.460 & 0.359 & 0.395 & \textbf{0.355} & \textbf{0.392}  \\

 & 720(720) & \textbf{0.526} & \textbf{0.497} & 0.589 & 0.550 & 0.415 & 0.423 & \textbf{0.413} & \textbf{0.421}  \\
 \midrule

\multirow{1}{*}{{$\bf 1^{st}$}} & count  & 2 & 4 & \textbf{18} & \textbf{16}  & 4 & 9 & \textbf{16} & \textbf{11}  \\

 \bottomrule
\end{tabular}
\caption{Results on base models based on CM. The forecasting horizon in parentheses is specific to Autoformer.}
\label{channel_mixing_c3rl}
\end{table}

\textbf{Quantitative analysis.}
We conduct a comprehensive evaluation of C3RL on nine datasets across seven base models and their C3RL-enhanced counterparts, using MSE and MAE as metrics. 
As shown in Table~\ref{channel_independence_c3rl} (full results shown in the \textbf{Appendix}), the base models enhanced by C3RL consistently demonstrate improved predictive performance. 
Specifically, S-Mamba achieves the best performance in only 25 out of 72 metrics (12 for MSE and 13 for MAE), making a best-score rate of 34.7\%, while S-Mamba (+C3RL) achieves 59 best scores (28 for MSE, 31 for MAE), raising the rate to 81.9\%.
DLinear obtains 31 best results (14 for MSE, 17 for MAE), with a rate of 43.1\%, whereas DLinear (+C3RL) improves to 52 best results (28 for MSE, 24 for MAE), reaching 72.2\%.
PatchTST achieves 24 best scores out of 56 metrics (13 for MSE, 11 for MAE), with a best-score rate of 42.9\%. In contrast, PatchTST (+C3RL) reaches 46 best scores (23 for MSE, 23 for MAE), achieving 82.1\%.
iTransformer obtains 30 best results (13 for MSE, 17 for MAE) with a 41.7\% rate, while iTransformer (+C3RL) achieves 59 best scores (32 for MSE, 27 for MAE), pushing the rate to 81.9\%.
RLinear reaches 40 best scores (20 for MSE, 20 for MAE), with a 55.6\% best-score rate. Enhanced with C3RL, it obtains 64 best scores (30 for MSE, 34 for MAE), increasing the rate to 88.9\%.
Overall, C3RL achieves an optimal rate of \textbf{81.4\%} across all tasks on the five models, representing a \textbf{37.8\%} improvement compared to the \textbf{43.6\%} achieved by models without C3RL. In addition, the results in Table \ref{channel_mixing_c3rl} show that applying C3RL to models based on CM strategies also yields strong performance (with the overall optimal rate increasing from \textbf{23.8\%} to \textbf{76.3\%}), and significant improvements are observed in many tasks. The selection of $\{\lambda_{simsia}, \lambda_{pred}\}$ is shown in the \textbf{Appendix}.

\begin{figure}[t]
	\centering
	\includegraphics[width=0.48\textwidth]{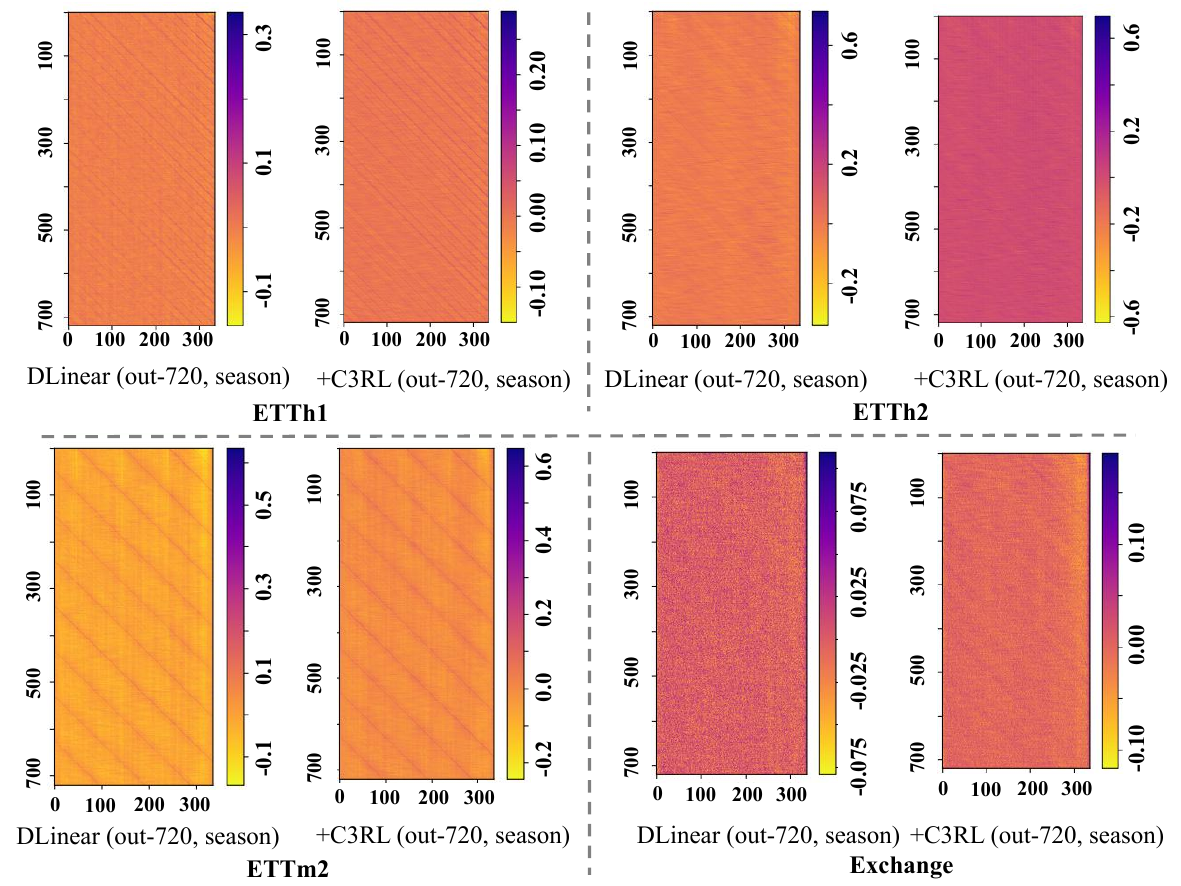}
	\caption{Visualization of the weights of DLinear on several datasets. Out-720 denotes that the prediction horizon is 720. Season means the seasonal item obtained by decomposition.}
	\label{weights_visualization}
\end{figure}

\textbf{Qualitative analysis.}
Figure \ref{weights_visualization} illustrates the seasonal weight visualizations of DLinear and DLinear (+C3RL). It is evident that DLinear (+C3RL) exhibits stronger structural clarity and greater discriminability in its visual representations, indicating enhanced feature modeling capability. For instance, in the ETTh2 and ETTm2 datasets, the weight heatmaps of DLinear (+C3RL) display overall deeper color intensities, suggesting significantly increased attention to specific time steps. Notably, in the ETTh1 and Exchange datasets, DLinear (+C3RL) demonstrates more distinct, periodically arranged dark band patterns, directly validating the effectiveness of C3RL in capturing seasonal patterns and improving representational expressiveness.
To further illustrate the impact of C3RL on prediction performance, we visualize the forecasting results of DLinear on two datasets in Figure \ref{prediction_visualization} (more results shown in the \textbf{Appendix}). As shown, incorporating C3RL enables the predicted curves to align more closely with the ground truth, indicating improved fitting accuracy. This is particularly prominent in the ETTh1 (out-192), where C3RL substantially enhances the model’s ability to capture complex temporal patterns, realizing more accurate forecasting of future trends. These results collectively validate the effectiveness of C3RL in improving the temporal representation capacity of the model.

\begin{figure}[t]
	\centering
	\includegraphics[width=0.45\textwidth]{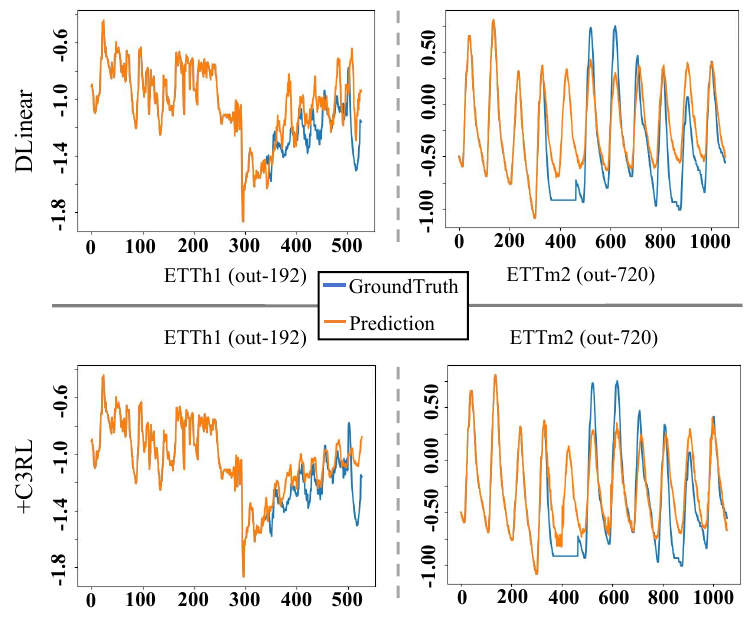}
	\caption{Visualization of the predictive performance.}
	\label{prediction_visualization}
\end{figure}

\subsection{Ablation Studies}

\begin{table*}[t]
\centering
\setlength{\tabcolsep}{1.3mm}
\scriptsize 
\begin{tabular}{l|r|cccc|cccc|cccc|cccc|cccc}
\toprule
\multicolumn{2}{c|}{Model} & \multicolumn{2}{c}{$\text{S-Mamba+}^{*}$} & \multicolumn{2}{c|}{\textbf{S-Mamba+}} & \multicolumn{2}{c}{$\text{DLinear+}^{*}$} & \multicolumn{2}{c|}{\textbf{DLinear+}} & \multicolumn{2}{c}{$\text{PatchTST+}^{*}$} & \multicolumn{2}{c|}{\textbf{PatchTST+}} & \multicolumn{2}{c}{$\text{iTrans+}^{*}$} & \multicolumn{2}{c|}{\textbf{iTrans+}} & \multicolumn{2}{c}{{$\text{RLinear+}^{*}$}} & \multicolumn{2}{c}{\textbf{RLinear+}} \\

\multicolumn{2}{c|}{Metric} & \multicolumn{1}{l}{MSE} &  {MAE} &  {MSE} &  {MAE} & \multicolumn{1}{l}{MSE} &  {MAE} &  {MSE} &  {MAE} & \multicolumn{1}{l}{MSE} &  {MAE} &  {MSE} &  {MAE} & \multicolumn{1}{l}{MSE} &  {MAE} & MSE & MAE & MSE & MAE & MSE & MAE\\

\midrule
\multirow{4}{*}{\rotatebox[origin=c]{90}{\textbf{ETTh1}}} & 96 & 0.402 & 0.420 & \textbf{0.386} & \textbf{0.405} & \textbf{0.374} & \textbf{0.395} & \textbf{0.374} & \textbf{0.395} & 0.379 & 0.403 & \textbf{0.376} & \textbf{0.400} & 0.396 & 0.414 & \textbf{0.387} & \textbf{0.405} & 0.383 & 0.406 & \textbf{0.367} & \textbf{0.391} \\

 & 192 & 0.449 & 0.447 & \textbf{0.442} & \textbf{0.442} & 0.409 & \textbf{0.418} & \textbf{0.408} & \textbf{0.418} & 0.415 & 0.422 & \textbf{0.414} & \textbf{0.421} & 0.446 & 0.441 & \textbf{0.441} & \textbf{0.438} & 0.413 & 0.420 & \textbf{0.404} & \textbf{0.412} \\
 
 & 336 & 0.490 & 0.468 & \textbf{0.485} & \textbf{0.465} & 0.465 & 0.464 & \textbf{0.443} & \textbf{0.444} & \textbf{0.429} & 0.435 & \textbf{0.429} & \textbf{0.434} & 0.491 & 0.465 & \textbf{0.485} & \textbf{0.461} & 0.428 & 0.435 & \textbf{0.418} & \textbf{0.423} \\
 
 & 720 & \textbf{0.501} & \textbf{0.494} & \textbf{0.501} & \textbf{0.494} & 0.544 & 0.537 & \textbf{0.472} & \textbf{0.489} & \textbf{0.444} & \textbf{0.462} & {0.445} & {0.463} & \textbf{0.503} & \textbf{0.491} & \textbf{0.503} & \textbf{0.491} & 0.437 & 0.454 & \textbf{0.431} & \textbf{0.449} \\
 
 \midrule
\multirow{4}{*}{\rotatebox[origin=c]{90}{\textbf{ETTh2}}} & 96 & 0.307 & 0.357 & \textbf{0.295} & \textbf{0.347} & 0.288 & 0.354 & \textbf{0.280} & \textbf{0.349} & 0.276 & 0.337 & \textbf{0.274} & \textbf{0.335} & 0.302 & 0.351 & \textbf{0.297} & \textbf{0.349} & 0.261 & \textbf{0.328} & \textbf{0.260} & \textbf{0.328} \\

 & 192 & 0.386 & 0.404 & \textbf{0.376} & \textbf{0.398} & 0.367 & 0.407 & \textbf{0.362} & \textbf{0.403} & 0.339 & \textbf{0.379} & \textbf{0.338} & \textbf{0.379} & 0.380 & 0.399 & \textbf{0.378} & \textbf{0.398} & 0.319 & 0.370 & \textbf{0.317} & \textbf{0.369} \\
 
 & 336 & 0.426 & 0.436 & \textbf{0.423} & \textbf{0.434} & 0.472 & 0.477 & \textbf{0.436} & \textbf{0.455} & \textbf{0.326} & \textbf{0.378} & {0.327} & \textbf{0.378} & 0.429 & 0.436 & \textbf{0.422} & \textbf{0.432} & 0.327 & 0.387 & \textbf{0.324} & \textbf{0.385} \\
 
 & 720 & 0.435 & 0.451 & \textbf{0.431} & \textbf{0.447} & 0.685 & 0.587 & \textbf{0.667} & \textbf{0.580} & \textbf{0.375} & \textbf{0.418} & \textbf{0.375} & \textbf{0.418} & 0.433 & 0.450 & \textbf{0.429} & \textbf{0.447} & 0.424 & 0.449 & \textbf{0.412} & \textbf{0.443} \\
 \midrule
 
\multirow{4}{*}{\rotatebox[origin=c]{90}{\textbf{ETTm2}}} & 96 & 0.186 & 0.271 & \textbf{0.179} & \textbf{0.264} & 0.172 & \textbf{0.268} & \textbf{0.171} & \textbf{0.268} & 0.170 & 0.258 & \textbf{0.163} & \textbf{0.252} & {0.184} & {0.270} & \textbf{0.183} & \textbf{0.267} & 0.165 & 0.254 & \textbf{0.164} & \textbf{0.252} \\

 & 192 & 0.255 & 0.315 & \textbf{0.248} & \textbf{0.309} & \textbf{0.235} & \textbf{0.317} & \textbf{0.235} & \textbf{0.317} & 0.226 & 0.297 & \textbf{0.220} & \textbf{0.292} & 0.254 & 0.314 & \textbf{0.251} & \textbf{0.312} & 0.223 & 0.294 & \textbf{0.219} & \textbf{0.290} \\
 
 & 336 & 0.317 & 0.353 & \textbf{0.311} & \textbf{0.348} & 0.318 & 0.379 & \textbf{0.313} & \textbf{0.373} & 0.281 & 0.333 & \textbf{0.277} & \textbf{0.330} & 0.317 & 0.353 & \textbf{0.315} & \textbf{0.351} & 0.278 & 0.330 & \textbf{0.274} & \textbf{0.326} \\
 
 & 720 & 0.416 & 0.407 & \textbf{0.413} & \textbf{0.405} & 0.393 & 0.412 & \textbf{0.392} & \textbf{0.411} & 0.364 & 0.383 & \textbf{0.362} & \textbf{0.382} & 0.413 & 0.406 & \textbf{0.410} & \textbf{0.405} & 0.369 & 0.385 & \textbf{0.367} & \textbf{0.384} \\
 \midrule

\multirow{4}{*}{\rotatebox[origin=c]{90}{\textbf{Exchange}}} & 96 & 0.092 & 0.215 & \textbf{0.088} & \textbf{0.208} & 0.085 & 0.208 & \textbf{0.082} & \textbf{0.202} & 0.099 & 0.222 & \textbf{0.093} & \textbf{0.213} & 0.092 & 0.215 & \textbf{0.087} & \textbf{0.208} & 0.095 & 0.219 & \textbf{0.090} & \textbf{0.209} \\

 & 192 & 0.188 & 0.310 & \textbf{0.180} & \textbf{0.302} & 0.165 & 0.298 & \textbf{0.160} & \textbf{0.291} & 0.196 & 0.317 & \textbf{0.194} & \textbf{0.314} & 0.186 & 0.308 & \textbf{0.178} & \textbf{0.302} & 0.195 & 0.317 & \textbf{0.186} & \textbf{0.305} \\
 
 & 336 & 0.343 & 0.426 & \textbf{0.331} & \textbf{0.417} & 0.365 & 0.456 & \textbf{0.319} & \textbf{0.435} & 0.356 & 0.435 & \textbf{0.352} & \textbf{0.433} & 0.352 & 0.432 & \textbf{0.338} & \textbf{0.422} & \textbf{0.357} & 0.435 & 0.361 & \textbf{0.434} \\
 
 & 720 & 0.877 & 0.710 & \textbf{0.862} & \textbf{0.701} & 1.022 & 0.766 & \textbf{0.888} & \textbf{0.721} & 0.906 & \textbf{0.713} & \textbf{0.904} & \textbf{0.713} & 0.870 & 0.705 & \textbf{0.852} & \textbf{0.697} & 0.923 & 0.720 & \textbf{0.919} & \textbf{0.719} \\

 \bottomrule
\end{tabular}
\caption{Performance comparison between weighted and unweighted base models enhanced by C3RL. ``+'' means the base models with C3RL and ``*'' denotes the models without weighting. iTrans is the iTransformer.
}
\label{without_weight}
\end{table*}

\begin{figure}[t]
	\centering
	\includegraphics[width=0.46\textwidth]{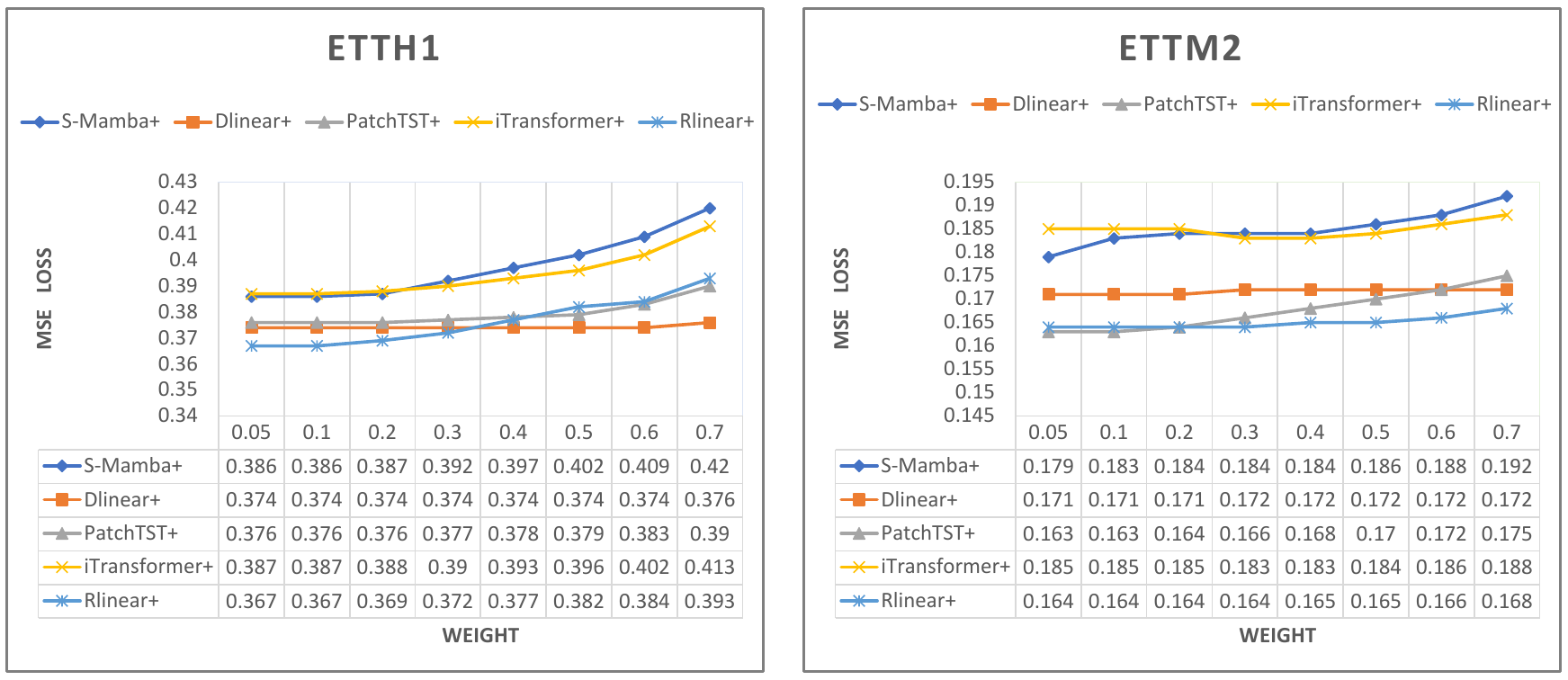}
	\caption{Influence of diverse WEIGHT ($\lambda_{simsia}$) on the performance of base models enhanced by C3RL.}
	\label{simsia_pred_weights}
\end{figure}

We conduct a comprehensive evaluation of five base models integrated with C3RL under various hyperparameter combinations $\{\lambda_{simsia}, \lambda_{pred}\}$, with the prediction length set to 96. The experimental results are illustrated in Figure~\ref{simsia_pred_weights}, where WEIGHT denotes $\lambda_{simsia}$ and the constraint $\lambda_{simsia} + \lambda_{pred} = 1$ holds. The symbol ``+'' indicates base models equipped with C3RL. As shown in the figure, increasing $\lambda_{simsia}$—i.e., placing more emphasis on representation learning during training—leads to a gradual rise in prediction error. This observation suggests that over-optimizing the representational capacity may compromise the model’s ability to perform downstream forecasting. Therefore, achieving a proper trade-off between representation learning and prediction objectives during joint training is crucial for enhancing overall model performance.

Furthermore, we set both $\lambda_{simsiam}$ and $\lambda_{pred}$ to 1 to evaluate the effectiveness of the proposed weighting mechanism. As shown in Table~\ref{without_weight}, it is evident that in nearly all tasks, the model incorporating C3RL with the weighting mechanism achieves significantly lower prediction errors compared to the baseline model that uses C3RL without weighting. These results demonstrate that appropriately balancing the contrastive and predictive losses not only preserves the representational capacity but also enhances performance on downstream forecasting tasks.

\section{Conclusion}
This paper proposes a novel MTSF framework based on representation learning, named C3RL, which integrates both CM and CI strategies. Inspired by the input formats required by these two strategies and the SimSiam representation learning paradigm, C3RL designs task-specific Siamese network structures tailored to the architectural characteristics of different base models—whether they adopt CM or CI strategy—by employing a complementary channel processing strategy. This design can effectively narrow the gap between feature representations learned from the same samples under different input formats, while mitigating the issue of collapsing solutions.
Moreover, to dynamically balance the objectives of representation learning and forecasting, C3RL introduces adjustable weighting parameters for the contrastive and predictive losses, thereby enhancing the base model’s representation capability while further improving its predictive performance. Extensive experimental results demonstrate the effectiveness of C3RL: it consistently improves forecasting accuracy across multiple datasets and strengthens the model's ability to perceive diverse feature structures. This work offers a novel paradigm for optimizing MTSF models, with strong potential for practical applications and broader adoption.

\bibliography{aaai2026}

\section{Appendix} 

\subsection{More instances with C3RL}

\subsubsection{S-Mamba+C3RL}

\begin{figure}[t]
	\centering
	\includegraphics[width=0.48\textwidth]{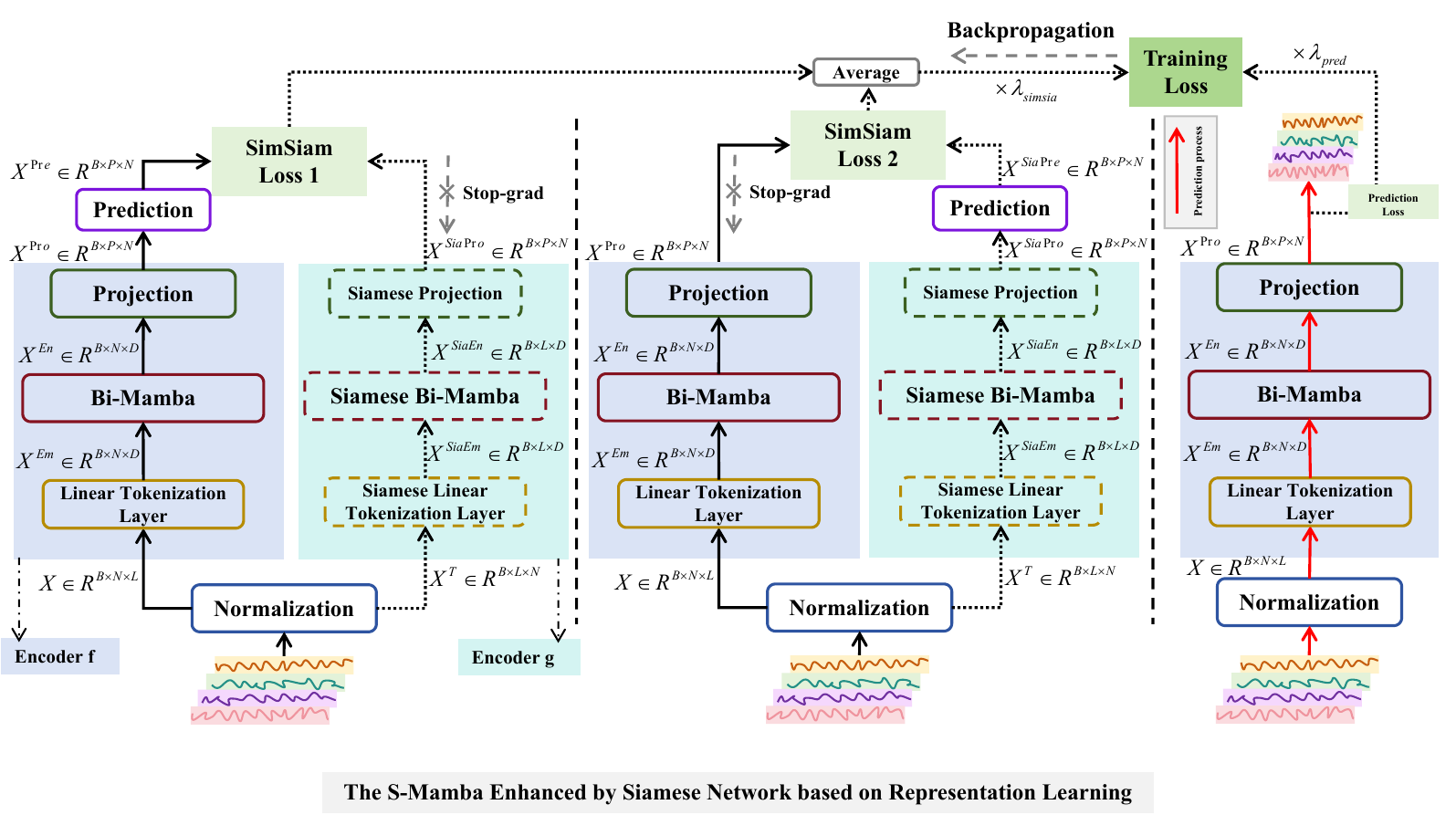}
	\caption{The application of C3RL to the S-Mamba.}
	\label{S-Mamba-c3rl}
\end{figure}

Figure~\ref{S-Mamba-c3rl} illustrates how C3RL is integrated into S-Mamba through the construction of a Siamese network:

Step 1: In S-Mamba, the Linear Tokenization Layer and Bi-Mamba modules are collectively referred to as T-Modules. These components can be directly adapted to construct the Siamese Linear Tokenization Layer and the Siamese Bi-Mamba.

Step 2: As the input dimensions differ on the Siamese side, the internal parameters of the Siamese Linear Tokenization Layer must be modified accordingly to ensure compatibility. Specifically, the original Linear Tokenization Layer is implemented as a linear layer with input feature dimension $L$, i.e., $nn.Linear(D_{in}, D)$ where $D_{in} = L$. In the Siamese variant, the input feature dimension becomes $N$, so $D_{in}$ should be updated to $N$.

Step 3: The output of the Siamese Linear Tokenization Layer, which feeds into the Siamese Bi-Mamba, retains the same last-dimensional size $D$ as in the original architecture. Therefore, the internal structure of the Siamese Bi-Mamba does not require modification.

\begin{figure}[t]
	\centering
	\includegraphics[width=0.3\textwidth]{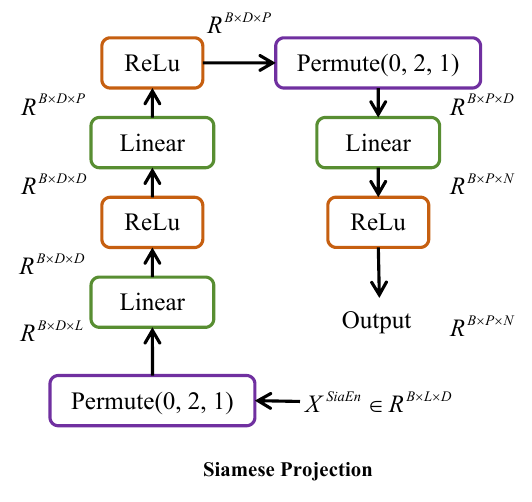}
	\caption{The structure of the S-Mamba's Siamese Projection.}
	\label{siamese-projection-S-Mamba}
\end{figure}

Step 4: The original Projection layer outputs $X^{Pro} \in \mathbb{R}^{B \times P \times N}$. To align the output of the Siamese Bi-Mamba, denoted as $X^{SiaEn} \in \mathbb{R}^{B \times L \times D}$, with $X^{Pro}$ in the feature space, a Siamese Projection layer is introduced, as illustrated in Figure~\ref{siamese-projection-S-Mamba}.

\subsubsection{DLinear+C3RL}

\begin{figure}[t]
	\centering
	\includegraphics[width=0.48\textwidth]{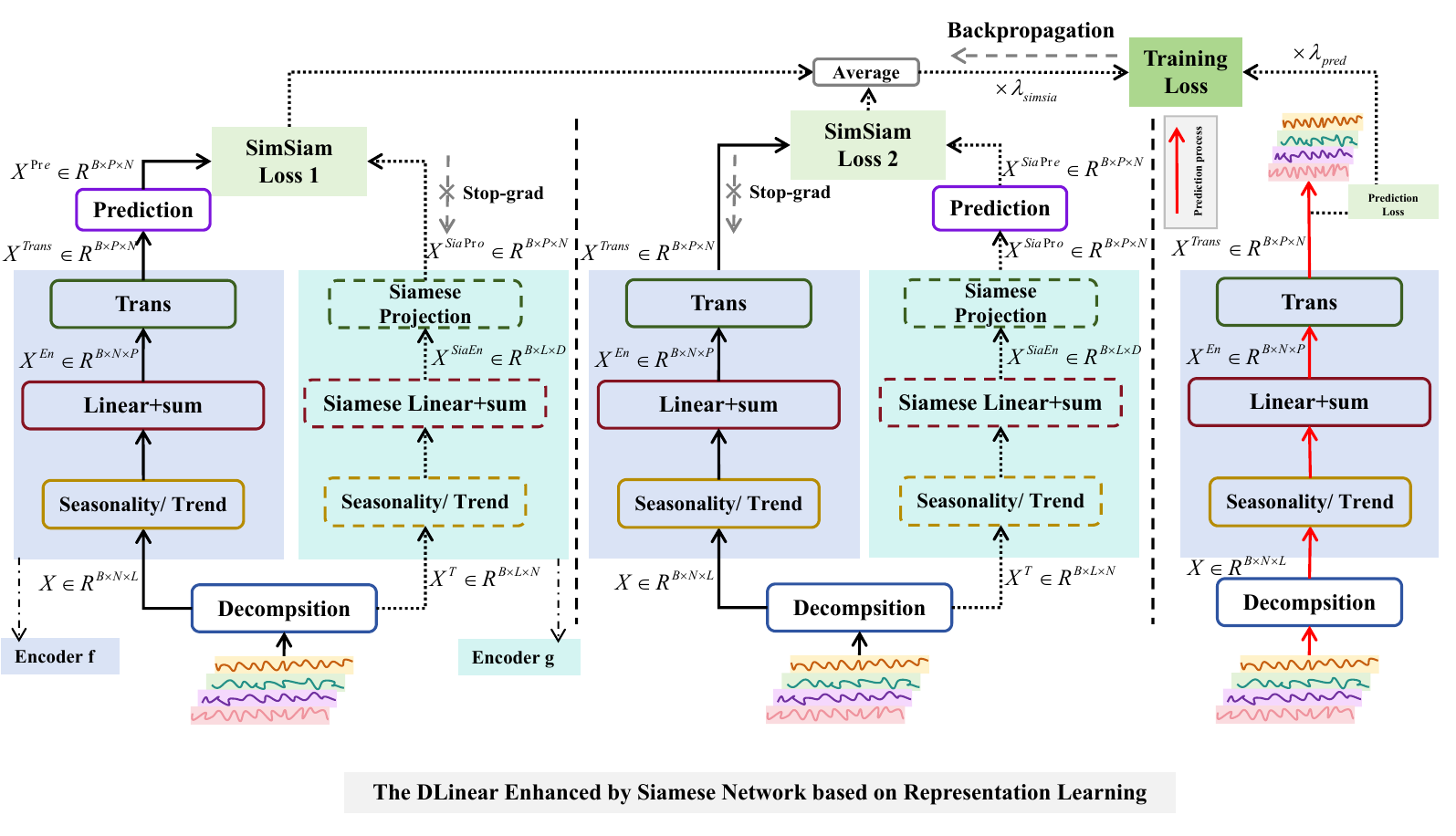}
	\caption{The application of C3RL to the DLinear.}
	\label{DLinear-c3rl}
\end{figure}

Figure~\ref{DLinear-c3rl} illustrates the construction of a Siamese network by integrating C3RL into DLinear.

Step 1: In DLinear, the Linear+sum layer functions as the T-Module and can be directly adapted to build the Siamese Linear+sum layer. Here, ``sum'' refers to the element-wise addition of the seasonality and trend components, each processed by their respective linear layers.

Step 2: Since the input dimensions differ on the Siamese branch, the internal parameters of the Siamese Linear layers must be adjusted to ensure correct data processing. Specifically, in the original model, the input feature dimension to the linear layer is $L$, i.e., defined as $nn.Linear(D_{in}, D)$ with $D_{in} = L$. For the Siamese branch, the input feature dimension becomes $N$, and thus $D_{in}$ is set to $N$ accordingly.

\begin{figure}[t]
	\centering
	\includegraphics[width=0.3\textwidth]{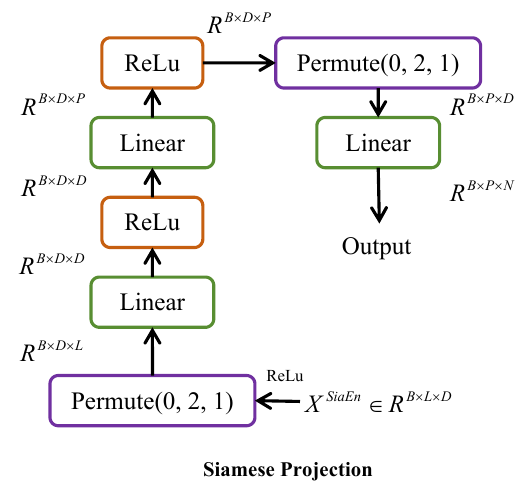}
	\caption{The structure of the DLinear's Siamese Projection.}
	\label{siamese-projection-DLinear}
\end{figure}

Step 3: The output of the original Trans layer is denoted as $X^{Trans} \in \mathbb{R}^{B \times P \times N}$, where ``Trans'' indicates a transposition operation. To align the Siamese Linear+sum output, $X^{SiaEn} \in \mathbb{R}^{B \times L \times D}$, with $X^{Trans}$ in the feature space, we introduce a Siamese Projection layer, as illustrated in Figure~\ref{siamese-projection-DLinear}.

\subsubsection{RLinear+C3RL}
\begin{figure}[t]
	\centering
	\includegraphics[width=0.48\textwidth]{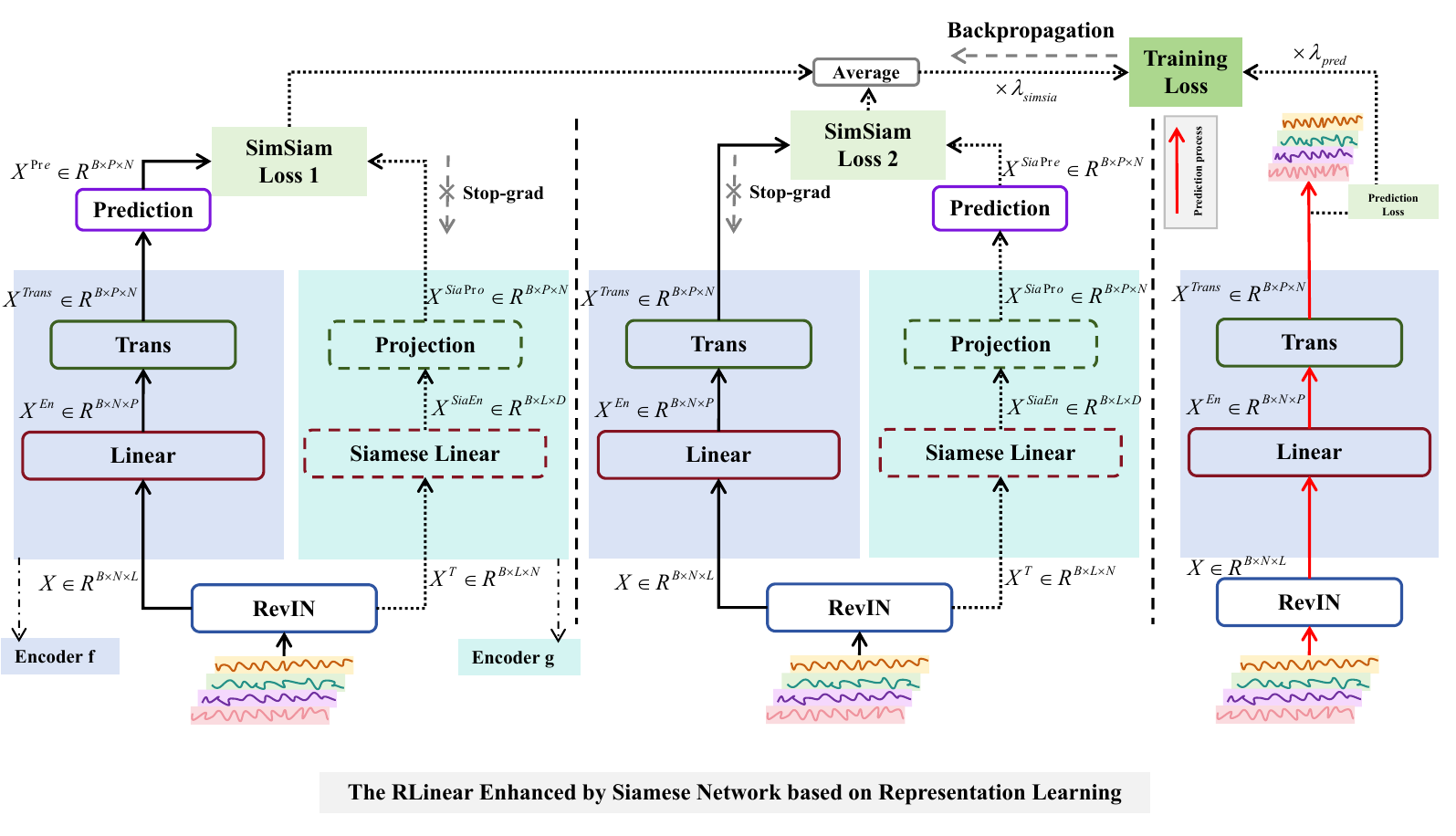}
	\caption{The application of C3RL to the RLinear.}
	\label{RLinear-c3rl}
\end{figure}

Figure~\ref{RLinear-c3rl} illustrates the construction of a Siamese network by integrating C3RL into RLinear, following modeling steps similar to those used in DLinear with C3RL.

\subsubsection{PatchTST+C3RL}

\begin{figure}[t]
	\centering
	\includegraphics[width=0.48\textwidth]{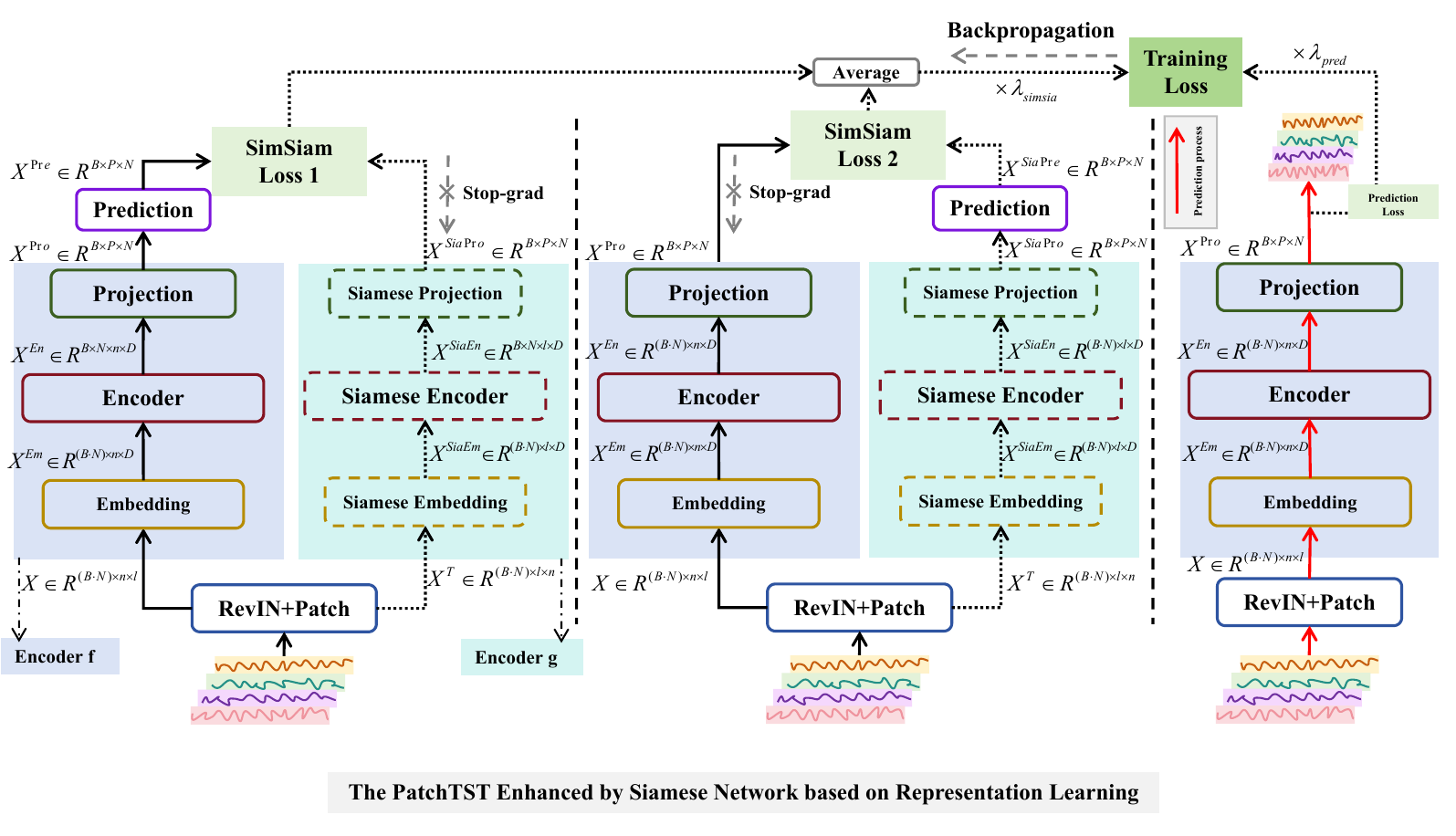}
	\caption{The application of C3RL to the PatchTST.}
	\label{PatchTST-c3rl}
\end{figure}

Figure~\ref{PatchTST-c3rl} illustrates how C3RL is integrated into PatchTST, which is the typical instance of the ECI strategy, to construct a Siamese network:

Step 1: In PatchTST, the Embedding and Encoder layers are regarded as part of the T-Modules. These components can be directly reused to build the Siamese Embedding and Siamese Encoder layers.

Step 2: Since the input dimensions differ on the Siamese side, the parameters of the Siamese Embedding layer must be adjusted to ensure compatibility. Specifically, the original Embedding layer (a linear layer) has input feature dimension $l$, i.e., $nn.Linear(D_{in}, D)$ with $D_{in} = l$. In the Siamese branch, the input dimension becomes $n$, thus requiring $D_{in} = n$ for proper alignment. Here, $l$ and $n$ denote the patch length and number of patches, respectively.

Step 3: The output of the Siamese Embedding layer, serving as the input to the Siamese Encoder, has the same last dimension $D$ as the original Encoder input, allowing the Siamese Encoder to retain its internal structure without modification.

Step 4: The original Projection layer outputs $X^{Pro} \in \mathbb{R}^{B \times P \times N}$. This is achieved by flattening $X^{En} \in \mathbb{R}^{B \times N \times n \times D}$ to $\mathbb{R}^{B \times N \times (n \cdot D)}$, applying a linear transformation to obtain $\mathbb{R}^{B \times N \times P}$, and finally transposing to the shape $\mathbb{R}^{B \times P \times N}$.
To align the output of the Siamese Encoder, $X^{SiaEn} \in \mathbb{R}^{B \times N \times l \times D}$, with the original projection output in feature space, we introduce a Siamese Projection layer. The key modification different from the original Projection layer lies in the linear transformation, which changes from $nn.Linear(n \cdot D, P)$ to $nn.Linear(l \cdot D, P)$.

\subsubsection{Informer+C3RL}

\begin{figure*}[t]
	\centering
	\includegraphics[width=\textwidth]{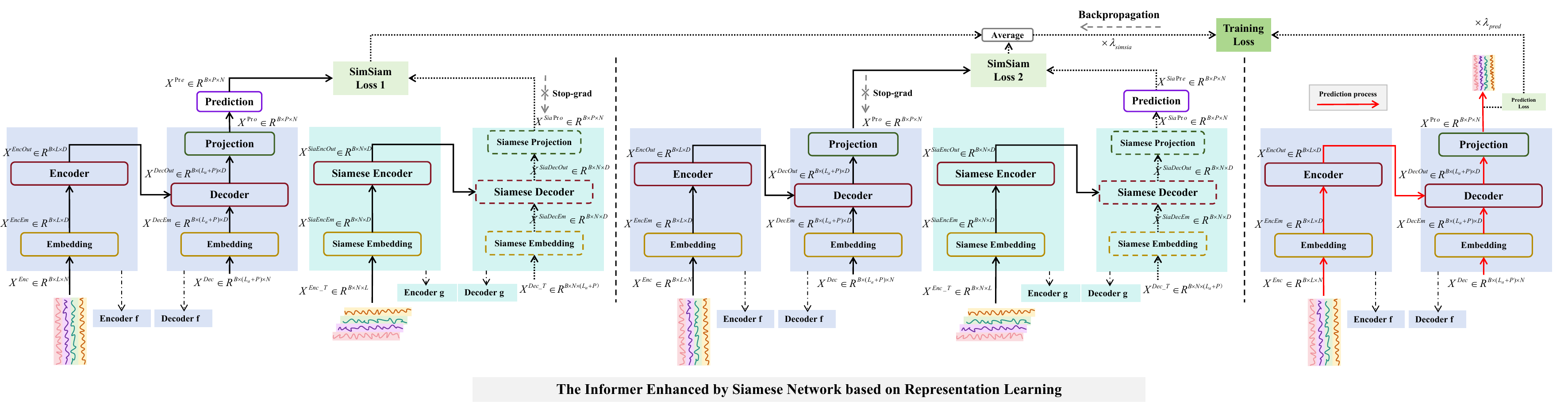}
	\caption{The application of C3RL to the Informer.}
	\label{Informer-c3rl}
\end{figure*}

Figure~\ref{Informer-c3rl} illustrates how C3RL is integrated into Informer through the construction of a Siamese network:

Step 1: In Informer, the $\text{Enc}\_\text{Embedding}$, $\text{Dec}\_\text{Embedding}$, Encoder and Decoder layers are collectively referred to as T-Modules. These components can be directly adapted to construct the Siamese $\text{Enc}\_\text{Embedding}$ layer, Siamese $\text{Dec}\_\text{Embedding}$ layer, Siamese Encoder layer and Siamese Decoder layer.

Step 2: As the input dimensions differ on the Siamese side, the internal parameters of the Siamese $\text{Enc}\_\text{Embedding}$ layer and Siamese $\text{Dec}\_\text{Embedding}$ layer must be modified accordingly to ensure compatibility. Specifically, the original $\text{Enc}\_\text{Embedding}$ layer can be implemented through a linear layer with input feature dimension $N$, i.e., $nn.Linear(D_{in}, D)$ where $D_{in} = N$. In the Siamese $\text{Enc}\_\text{Embedding}$, the input feature dimension becomes $L$, so $D_{in}$ should be updated to $L$.
Similarly, the original $\text{Dec}\_\text{Embedding}$ layer can be implemented through a linear layer with input feature dimension $N$, i.e., $nn.Linear(D_{in}, D)$ where $D_{in} = N$. In the Siamese $\text{Dec}\_\text{Embedding}$, the input feature dimension becomes $L_a + P$, so $D_{in}$ should be updated to $L_a + P$, where $L_a$ denotes the length of partial historical values and $P$ represents the forecasting horizon.

Step 3: The output of Siamese $\text{Enc}\_\text{Embedding}$ layer and Siamese $\text{Dec}\_\text{Embedding}$ layer, which feeds into the Siamese Encoder and Siamese Decoder respectively, retains the same last-dimensional size $D$ as in the original architecture. Therefore, the internal structure of Siamese Encoder and Siamese Decoder does not require modification.

\begin{figure}[t]
	\centering
	\includegraphics[width=0.2\textwidth]{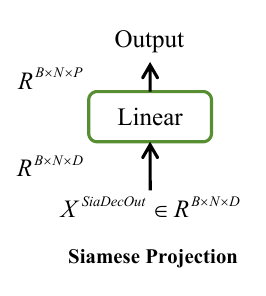}
	\caption{The structure of the Informer's Siamese Projection.}
	\label{siamese-projection-Informer}
\end{figure}

Step 4: The original Projection layer outputs $X^{Pro} \in \mathbb{R}^{B \times P \times N}$. To align the output of the Siamese Decoder, denoted as $X^{SiaDecOut} \in \mathbb{R}^{B \times N \times D}$, with $X^{Pro}$ in the feature space, a Siamese Projection layer is introduced, as illustrated in Figure~\ref{siamese-projection-Informer}.

\subsubsection{Autoformer+C3RL}
The application of C3RL to the Autoformer is similar to the Informer+C3RL.

\subsection{Dataset Description}
ETT includes seven variables related to electricity transformers, collected from July 2016 to July 2018. It comprises four subsets: ETTh1 and ETTh2 are sampled hourly, while ETTm1 and ETTm2 are sampled every 15 minutes.
The Electricity dataset contains the hourly electricity consumption of 321 customers from 2012 to 2014. 
The Weather dataset is collected every 10 minutes from the Weather Station of the Max Planck Biogeochemistry Institute in 2020, containing 21 meteorological indicators, such as humidity, air temperature, etc. 
The Traffic dataset consists of hourly road occupancy data from 862 sensors on San Francisco Bay area freeways, provided by the California Department of Transportation, covering the period from January 2015 to December 2016. 
The Exchange dataset gathers daily exchange rates for eight countries from 1990 to 2016. 
The Illness dataset contains weekly records of influenza-like illness (ILI) cases collected by the U.S. Centers for Disease Control and Prevention from 2002 to 2021, representing the proportion of ILI patients relative to the total number of clinical visits.

\subsection{Evaluation Metrics}

We adopt Mean Absolute Error (MAE) and Mean Squared Error (MSE) to assess model performance, calculated as:
\begin{equation}
\begin{split}
    & M A E=\frac{1}{\tau} \sum_{i=t_0}^{t_0+\tau-1}\left|\hat{x}_i-x_i\right| \\
	& M S E=\frac{1}{\tau} \sum_{i=t_0}^{t_0+\tau-1}\left(\hat{x}_i-x_i\right)^2 ,
\end{split}
\end{equation}
where $\hat{x_i}$ is the predicted value, $x_i$ is the ground truth, and $\tau$ denotes the prediction horizon. Smaller MAE and MSE values indicate higher forecasting accuracy.

\begin{figure*}[t]
	\centering
	\includegraphics[width=\textwidth]{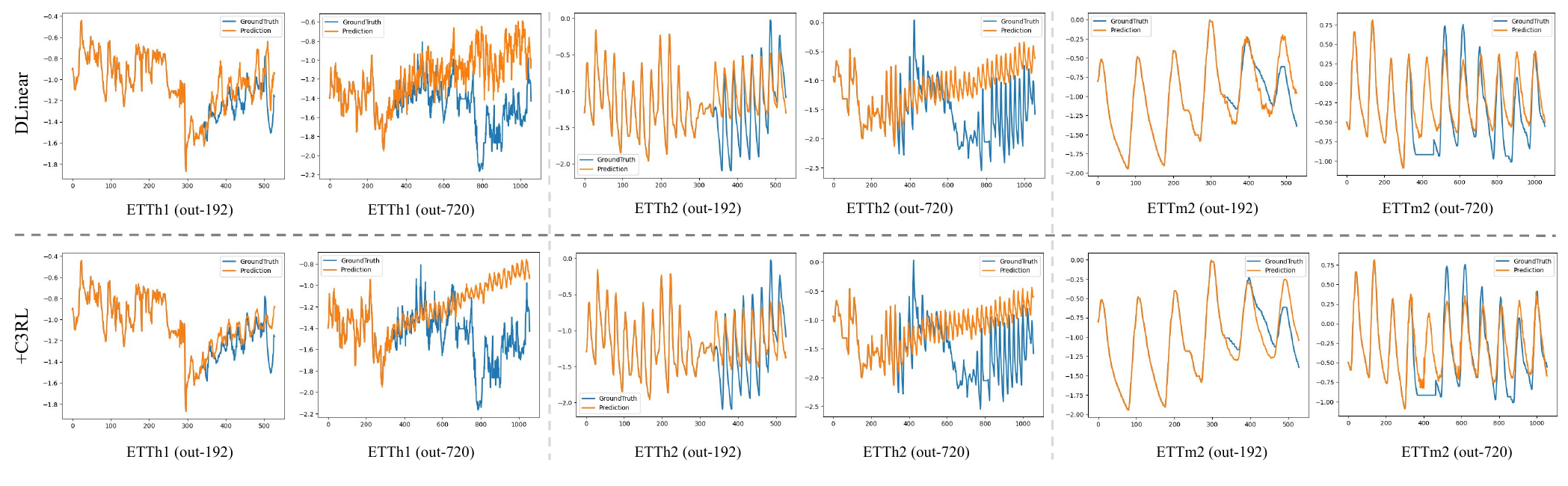}
	\caption{More visualization of the predictive performance.}
	\label{prediction_visualization_full}
\end{figure*}

\subsection{Baselines}
S-Mamba applies a linear projection to encode time steps for each variable separately, then leverages a bidirectional Mamba module to model inter-variable relationships, and captures temporal patterns using a feed-forward structure.

DLinear introduces the earliest linear approach in multivariate time series forecasting, adopting a simple one-layer architecture based on a decomposition method.

PatchTST divides time series data into smaller segments (patches), handling each variable independently while still effectively leveraging extended historical sequences.

iTransformer modifies the conventional Transformer input format by regarding the temporal sequence of each feature as an individual token, and utilizes self-attention to learn variable-level interactions.

RLinear revisits the function of linear transformations in MTSF, and investigates how reversible normalization and the principle of channel-wise independence contribute to improved predictive accuracy.

Informer introduces the ProbSparse self-attention and self-attention distilling mechanisms to capture the most critical keys and reduce computation overhead.

Autoformer designs a decomposition-based framework to extract more predictable components from complex temporal trends and leverages an auto-correlation mechanism to model dependencies across entire time series.

\subsection{Full Experiment Results}
The complete results on CI-based models are shown in Table~\ref{channel_independence_c3rl_full}. The selection of $\{\lambda_{simsia}, \lambda_{pred} \}$ for CI-based and CM-based models can be found in Table~\ref{lambda_selection_CI} and Table~\ref{lambda_selection_CM}, respectively. More visualization of the predictive performance can be seen in Figure \ref{prediction_visualization_full}.

\begin{table*}[ht]
\centering
\setlength{\tabcolsep}{1.3mm}
\scriptsize 
\begin{tabular}{l|r|cccc|cccc|cccc|cccc|cccc}
\toprule
\multicolumn{2}{c|}{Model} & \multicolumn{2}{c}{S-Mamba} & \multicolumn{2}{c|}{\textbf{+ C3RL}} & \multicolumn{2}{c}{DLinear} & \multicolumn{2}{c|}{\textbf{+ C3RL}} & \multicolumn{2}{c}{PatchTST} & \multicolumn{2}{c|}{\textbf{+ C3RL}} & \multicolumn{2}{c}{iTransformer} & \multicolumn{2}{c|}{\textbf{+ C3RL}} & \multicolumn{2}{c}{{RLinear}} & \multicolumn{2}{c}{\textbf{+ C3RL}} \\

\multicolumn{2}{c|}{Metric} & \multicolumn{1}{l}{MSE} &  {MAE} &  {MSE} &  {MAE} & \multicolumn{1}{l}{MSE} &  {MAE} &  {MSE} &  {MAE} & \multicolumn{1}{l}{MSE} &  {MAE} &  {MSE} &  {MAE} & \multicolumn{1}{l}{MSE} &  {MAE} & MSE & MAE & MSE & MAE & MSE & MAE\\

\midrule
\multirow{4}{*}{\rotatebox[origin=c]{90}{\textbf{ETTh1}}} & 96 & 0.388 & 0.406 & \textbf{0.386} & \textbf{0.405} & 0.384 & 0.405 & \textbf{0.374} & \textbf{0.395} & \textbf{0.375} & \textbf{0.399} & 0.376 & 0.400 & \textbf{0.387} & \textbf{0.405} & \textbf{0.387} & \textbf{0.405} & \textbf{0.366} & \textbf{0.391} & 0.367 & \textbf{0.391} \\

 & 192 & 0.445 & \textbf{0.441} & \textbf{0.442} & 0.442 & 0.443 & 0.450 & \textbf{0.408} & \textbf{0.418} & \textbf{0.414} & \textbf{0.421} & \textbf{0.414} & \textbf{0.421} & \textbf{0.441} & \textbf{0.436} & \textbf{0.441} & 0.438 & \textbf{0.403} & \textbf{0.412} & 0.404 & \textbf{0.412} \\
 
 & 336 & 0.490 & \textbf{0.465} & \textbf{0.485} & \textbf{0.465} & 0.447 & 0.448 & \textbf{0.443} & \textbf{0.444} & 0.431 & 0.435 & \textbf{0.429} & \textbf{0.434} & 0.491 & 0.462 & \textbf{0.485} & \textbf{0.461} & 0.420 & \textbf{0.423} & \textbf{0.418} & \textbf{0.423} \\
 
 & 720 & 0.506 & 0.497 & \textbf{0.501} & \textbf{0.494} & 0.504 & 0.515 & \textbf{0.472} & \textbf{0.489} & 0.450 & 0.466 & \textbf{0.445} & \textbf{0.463} & 0.509 & 0.494 & \textbf{0.503} & \textbf{0.491} & 0.442 & 0.456 & \textbf{0.431} & \textbf{0.449} \\
 
 \midrule
\multirow{4}{*}{\rotatebox[origin=c]{90}{\textbf{ETTh2}}} & 96 & 0.297 & 0.349 & \textbf{0.295} & \textbf{0.347} & 0.290 & 0.353 & \textbf{0.280} & \textbf{0.349} & \textbf{0.274} & 0.336 & \textbf{0.274} & \textbf{0.335} & 0.301 & 0.350 & \textbf{0.297} & \textbf{0.349} & 0.262 & 0.331 & \textbf{0.260} & \textbf{0.328} \\

 & 192 & 0.378 & 0.399 & \textbf{0.376} & \textbf{0.398} & 0.388 & 0.422 & \textbf{0.362} & \textbf{0.403} & \textbf{0.338} & \textbf{0.378} & \textbf{0.338} & 0.379 & 0.380 & 0.399 & \textbf{0.378} & \textbf{0.398} & 0.320 & 0.374 & \textbf{0.317} & \textbf{0.369} \\
 
 & 336 & 0.425 & 0.435 & \textbf{0.423} & \textbf{0.434} & 0.463 & 0.473 & \textbf{0.436} & \textbf{0.455} & 0.331 & 0.380 & \textbf{0.327} & \textbf{0.378} & 0.424 & \textbf{0.432} & \textbf{0.422} & \textbf{0.432} & 0.326 & 0.388 & \textbf{0.324} & \textbf{0.385} \\
 
 & 720 & 0.432 & 0.448 & \textbf{0.431} & \textbf{0.447} & 0.733 & 0.606 & \textbf{0.667} & \textbf{0.580} & 0.379 & 0.421 & \textbf{0.375} & \textbf{0.418} & 0.430 & \textbf{0.447} & \textbf{0.429} & \textbf{0.447} & 0.425 & 0.449 & \textbf{0.412} & \textbf{0.443} \\
 \midrule
 
\multirow{4}{*}{\rotatebox[origin=c]{90}{\textbf{ETTm1}}} & 96 & \textbf{0.331} & \textbf{0.368} & 0.332 & 0.369 & 0.301 & 0.345 & \textbf{0.300} & \textbf{0.344} & 0.292 & 0.343 & \textbf{0.289} & \textbf{0.342} & \textbf{0.342} & 0.377 & \textbf{0.342} & \textbf{0.376} & \textbf{0.301} & \textbf{0.343} & 0.302 & \textbf{0.343} \\

 & 192 & 0.378 & \textbf{0.393} & \textbf{0.377} & \textbf{0.393} & \textbf{0.336} & \textbf{0.366} & 0.337 & 0.367 & \textbf{0.331} & \textbf{0.369} & 0.332 & \textbf{0.369} & 0.383 & 0.396 & \textbf{0.381} & \textbf{0.394} & 0.341 & 0.367 & \textbf{0.338} & \textbf{0.365} \\
 
 & 336 & 0.410 & \textbf{0.414} & \textbf{0.408} & \textbf{0.414} & \textbf{0.372} & \textbf{0.389} & 0.375 & 0.391 & \textbf{0.365} & 0.392 & 0.366 & \textbf{0.391} & 0.418 & 0.418 & \textbf{0.416} & \textbf{0.417} & 0.374 & 0.386 & \textbf{0.371} & \textbf{0.384} \\
 
 & 720 & \textbf{0.474} & \textbf{0.451} & 0.475 & 0.453 & \textbf{0.427} & \textbf{0.423} & 0.432 & 0.427 & 0.421 & 0.425 & \textbf{0.420} & \textbf{0.424} & 0.487 & 0.457 & \textbf{0.484} & \textbf{0.454} & 0.430 & 0.418 & \textbf{0.426} & \textbf{0.415} \\
 \midrule
 
\multirow{4}{*}{\rotatebox[origin=c]{90}{\textbf{ETTm2}}} & 96 & 0.182 & 0.266 & \textbf{0.179} & \textbf{0.264} & 0.172 & \textbf{0.267} & \textbf{0.171} & 0.268 & 0.164 & 0.254 & \textbf{0.163} & \textbf{0.252} & 0.186 & 0.272 & \textbf{0.183} & \textbf{0.267} & \textbf{0.164} & 0.253 & \textbf{0.164} & \textbf{0.252} \\

 & 192 & 0.252 & 0.313 & \textbf{0.248} & \textbf{0.309} & 0.237 & \textbf{0.314} & \textbf{0.235} & 0.317 & 0.221 & \textbf{0.292} & \textbf{0.220} & \textbf{0.292} & 0.254 & 0.314 & \textbf{0.251} & \textbf{0.312} & \textbf{0.219} & \textbf{0.290} & \textbf{0.219} & \textbf{0.290} \\
 
 & 336 & 0.313 & 0.349 & \textbf{0.311} & \textbf{0.348} & \textbf{0.295} & \textbf{0.359} & 0.313 & 0.373 & 0.278 & \textbf{0.329} & \textbf{0.277} & 0.330 & 0.316 & \textbf{0.351} & \textbf{0.315} & \textbf{0.351} & \textbf{0.273} & \textbf{0.326} & 0.274 & \textbf{0.326} \\
 
 & 720 & \textbf{0.413} & \textbf{0.405} & \textbf{0.413} & \textbf{0.405} & 0.427 & 0.439 & \textbf{0.392} & \textbf{0.411} & 0.367 & 0.385 & \textbf{0.362} & \textbf{0.382} & 0.414 & 0.407 & \textbf{0.410} & \textbf{0.405} & \textbf{0.366} & 0.385 & 0.367 & \textbf{0.384} \\
 \midrule
 
\multirow{4}{*}{\rotatebox[origin=c]{90}{\textbf{Exchange}}} & 96 & \textbf{0.086} & \textbf{0.206} & 0.088 & 0.208 & 0.085 & 0.209 & \textbf{0.082} & \textbf{0.202} & \textbf{0.093} & \textbf{0.213} & \textbf{0.093} & \textbf{0.213} & \textbf{0.086} & \textbf{0.206} & 0.087 & 0.208 & \textbf{0.090} & \textbf{0.209} & \textbf{0.090} & \textbf{0.209} \\

 & 192 & 0.182 & 0.304 & \textbf{0.180} & \textbf{0.302} & 0.162 & 0.296 & \textbf{0.160} & \textbf{0.291} & \textbf{0.194} & \textbf{0.314} & \textbf{0.194} & \textbf{0.314} & 0.181 & 0.304 & \textbf{0.178} & \textbf{0.302} & 0.193 & 0.311 & \textbf{0.186} & \textbf{0.305} \\
 
 & 336 & \textbf{0.331} & \textbf{0.417} & \textbf{0.331} & \textbf{0.417} & 0.333 & 0.441 & \textbf{0.319} & \textbf{0.435} & 0.355 & 0.436 & \textbf{0.352} & \textbf{0.433} & \textbf{0.338} & \textbf{0.422} & \textbf{0.338} & \textbf{0.422} & 0.362 & 0.435 & \textbf{0.361} & \textbf{0.434} \\
 
 & 720 & \textbf{0.858} & \textbf{0.699} & 0.862 & 0.701 & 0.898 & 0.725 & \textbf{0.888} & \textbf{0.721} & \textbf{0.903} & \textbf{0.712} & 0.904 & 0.713 & 0.853 & \textbf{0.696} & \textbf{0.852} & 0.697 & 0.923 & \textbf{0.719} & \textbf{0.919} & \textbf{0.719} \\
 \midrule
 
\multirow{4}{*}{\rotatebox[origin=c]{90}{\textbf{Weather}}} & 96 & \textbf{0.165} & \textbf{0.209} & 0.166 & \textbf{0.209} & \textbf{0.174} & \textbf{0.233} & 0.176 & 0.238 & \textbf{0.150} & \textbf{0.198} & \textbf{0.150} & \textbf{0.198} & 0.176 & \textbf{0.216} & \textbf{0.175} & 0.217 & \textbf{0.175} & \textbf{0.225} & \textbf{0.175} & \textbf{0.225} \\

 & 192 & \textbf{0.215} & 0.255 & \textbf{0.215} & \textbf{0.254} & 0.218 & 0.278 & \textbf{0.216} & \textbf{0.273} & \textbf{0.195} & 0.242 & \textbf{0.195} & \textbf{0.241} & 0.225 & \textbf{0.257} & \textbf{0.223} & 0.258 & \textbf{0.217} & \textbf{0.259} & 0.218 & 0.260 \\
 
 & 336 & \textbf{0.273} & 0.296 & \textbf{0.273} & \textbf{0.295} & 0.263 & 0.314 & \textbf{0.261} & \textbf{0.311} & 0.250 & 0.284 & \textbf{0.247} & \textbf{0.283} & \textbf{0.281} & 0.299 & \textbf{0.281} & \textbf{0.298} & \textbf{0.265} & \textbf{0.294} & \textbf{0.265} & \textbf{0.294} \\
 
 & 720 & \textbf{0.353} & 0.349 & \textbf{0.353} & \textbf{0.348} & 0.332 & 0.374 & \textbf{0.328} & \textbf{0.367} & \textbf{0.320} & 0.336 & \textbf{0.320} & \textbf{0.335} & \textbf{0.358} & 0.350 & \textbf{0.358} & \textbf{0.349} & \textbf{0.329} & \textbf{0.339} & \textbf{0.329} & \textbf{0.339} \\ 
 \midrule
 
\multirow{4}{*}{\rotatebox[origin=c]{90}{\textbf{Electricity}}} & 96 & 0.140 & 0.236 & \textbf{0.137} & \textbf{0.235} & \textbf{0.140} & \textbf{0.237} & \textbf{0.140} & \textbf{0.237} & N/A & N/A & N/A & N/A & \textbf{0.148} & \textbf{0.239} & \textbf{0.148} & 0.240 & \textbf{0.141} & \textbf{0.236} & \textbf{0.141} & \textbf{0.236} \\

 & 192 & 0.161 & 0.258 & \textbf{0.158} & \textbf{0.255} & \textbf{0.154} & \textbf{0.250} & \textbf{0.154} & \textbf{0.250} & N/A & N/A & N/A & N/A & 0.166 & 0.258 & \textbf{0.164} & \textbf{0.255} & \textbf{0.154} & \textbf{0.248} & \textbf{0.154} & \textbf{0.248} \\
 
 & 336 & 0.180 & 0.277 & \textbf{0.177} & \textbf{0.276} & \textbf{0.169} & \textbf{0.268} & \textbf{0.169} & \textbf{0.268} & N/A & N/A & N/A & N/A & \textbf{0.178} & \textbf{0.271} & \textbf{0.178} & \textbf{0.271} & \textbf{0.171} & \textbf{0.264} & \textbf{0.171} & \textbf{0.264} \\
 
 & 720 & 0.201 & \textbf{0.297} & \textbf{0.199} & \textbf{0.297} & \textbf{0.204} & \textbf{0.300} & \textbf{0.204} & 0.301 & N/A & N/A & N/A & N/A & 0.211 & \textbf{0.299} & \textbf{0.210} & 0.300 & \textbf{0.210} & \textbf{0.297} & \textbf{0.210} & \textbf{0.297} \\ 
 \midrule

 \multirow{4}{*}{\rotatebox[origin=c]{90}{\textbf{Traffic}}} & 96 & 0.385 & 0.261 & \textbf{0.382} & \textbf{0.258} & 0.413 & 0.287 & \textbf{0.412} & \textbf{0.286} & N/A & N/A & N/A & N/A & \textbf{0.392} & \textbf{0.268} & 0.393 & 0.269 & \textbf{0.409} & \textbf{0.281} & \textbf{0.409} & 0.282 \\

 & 192 & 0.404 & 0.271 & \textbf{0.394} & \textbf{0.266} & \textbf{0.424} & \textbf{0.290} & \textbf{0.424} & \textbf{0.290} & N/A & N/A & N/A & N/A & 0.413 & \textbf{0.277} & \textbf{0.410} & \textbf{0.277} & \textbf{0.414} & \textbf{0.282} & \textbf{0.414} & \textbf{0.282} \\
 
 & 336 & \textbf{0.416} & 0.280 & 0.417 & \textbf{0.279} & \textbf{0.438} & \textbf{0.299} & \textbf{0.438} & \textbf{0.299} & N/A & N/A & N/A & N/A & \textbf{0.424} & \textbf{0.283} & \textbf{0.424} & \textbf{0.283} & \textbf{0.428} & \textbf{0.290} & \textbf{0.428} & \textbf{0.290} \\
 
 & 720 & 0.462 & 0.298 & \textbf{0.459} & \textbf{0.296} & \textbf{0.466} & \textbf{0.316} & 0.468 & 0.318 & N/A & N/A & N/A & N/A & \textbf{0.458} & \textbf{0.300} & 0.459 & 0.301 & \textbf{0.463} & \textbf{0.310} & \textbf{0.463} & \textbf{0.310} \\
 \midrule
 
\multirow{4}{*}{\rotatebox[origin=c]{90}{\textbf{Illness}}} & 24 & 1.849 & 0.865 & \textbf{1.808} & \textbf{0.854} & \textbf{2.280} & \textbf{1.061} & 2.289 & 1.064 & \textbf{1.401} & \textbf{0.738} & 1.522 & 0.815 & 2.358 & 1.063 & \textbf{2.345} & \textbf{1.051} & 2.306 & 1.037 & \textbf{2.303} & \textbf{1.032} \\

 & 36 & 2.188 & 1.006 & \textbf{1.776} & \textbf{0.877} & \textbf{2.235} & \textbf{1.059} & 2.274 & 1.063 & 1.452 & 0.839 & \textbf{1.317} & \textbf{0.781} & 2.184 & 1.002 & \textbf{2.182} & \textbf{0.999} & 2.173 & 1.014 & \textbf{2.069} & \textbf{0.977} \\
 
 & 48 & \textbf{1.708} & \textbf{0.875} & 1.787 & 0.894 & 2.298 & \textbf{1.079} & \textbf{2.285} & 1.087 & 1.679 & 0.859 & \textbf{1.498} & \textbf{0.817} & \textbf{2.124} & 1.019 & 2.161 & \textbf{1.018} & 2.184 & 1.005 & \textbf{2.151} & \textbf{0.996} \\
 
 & 60 & 2.338 & 1.005 & \textbf{2.068} & \textbf{0.953} & 2.573 & 1.157 & \textbf{2.420} & \textbf{1.111} & 1.621 & 0.888 & \textbf{1.395} & \textbf{0.798} & 2.165 & 1.032 & \textbf{2.061} & \textbf{0.991} & 2.087 & 0.983 & \textbf{2.080} & \textbf{0.981} \\
  \midrule

\multirow{1}{*}{{$\bf 1^{st}$}} & count  & 12 & 13 & \textbf{29} & \textbf{30}  & 14 & 17 & \textbf{28} & \textbf{24}  & 13 & 11 & \textbf{23} & \textbf{23}  & 13 & 17 & \textbf{32} & \textbf{27}  & 20 & 20 & \textbf{30} & \textbf{34} \\
 \bottomrule
\end{tabular}
\caption{Results on models based on CI. The better performance in each setting is shown in \textbf{bold}. N/A means out of memory.
}
\label{channel_independence_c3rl_full}
\end{table*}

\begin{table*}[ht]
\centering
\begin{tabular}{l|r|cc|cc|cc|cc|cc}
\toprule
\multicolumn{2}{c|}{Model} & \multicolumn{2}{c|}{S-Mamba+} & \multicolumn{2}{c|}{DLinear+} & \multicolumn{2}{c|}{PatchTST+} & \multicolumn{2}{c|}{iTransformer+} & \multicolumn{2}{c}{{RLinear+}} \\

\multicolumn{2}{c|}{Parameter} & \multicolumn{1}{l}{$\lambda_{simsia}$} &  {$\lambda_{pred}$} & \multicolumn{1}{l}{$\lambda_{simsia}$} &  {$\lambda_{pred}$} & \multicolumn{1}{l}{$\lambda_{simsia}$} &  {$\lambda_{pred}$} & \multicolumn{1}{l}{$\lambda_{simsia}$} &  {$\lambda_{pred}$} & $\lambda_{simsia}$ & $\lambda_{pred}$ \\

\midrule
\multirow{4}{*}{\rotatebox[origin=c]{90}{\textbf{ETTh1}}} & 96 & 0.1 & 0.9 & 0.4 & 0.6 & 0.05 & 0.95 & 0.1 & 0.9 & 0.05 & 0.95 \\

 & 192 & 0.2 & 0.8 & 0.3 & 0.7 & 0.05 & 0.95 & 0.3 & 0.7 & 0.01 & 0.99 \\
 
 & 336 & 0.3 & 0.7 & 0.4 & 0.6 & 0.4 & 0.6 & 0.4 & 0.6 & 0.1 & 0.9 \\
 
 & 720 & 0.5 & 0.5 & 0.5 & 0.5 & 0.4 & 0.6 & 0.5 & 0.5 & 0.2 & 0.8 \\
 
 \midrule
\multirow{4}{*}{\rotatebox[origin=c]{90}{\textbf{ETTh2}}} & 96 & 0.1 & 0.9 & 0.4 & 0.6 & 0.2 & 0.8 & 0.2 & 0.8 & 0.2 & 0.8 \\

 & 192 & 0.2 & 0.8 & 0.4 & 0.6 & 0.2 & 0.8 & 0.2 & 0.8 & 0.2 & 0.8 \\
 
 & 336 & 0.2 & 0.8 & 0.4 & 0.6 & 0.4 & 0.6 & 0.2 & 0.8 & 0.2 & 0.8 \\
 
 & 720 & 0.05 & 0.95 & 0.4 & 0.6 & 0.4 & 0.6 & 0.1 & 0.9 & 0.2 & 0.8 \\
 \midrule
 
\multirow{4}{*}{\rotatebox[origin=c]{90}{\textbf{ETTm1}}} & 96 & 0.05 & 0.95 & 0.4 & 0.6 & 0.2 & 0.8 & 0.2 & 0.8 & 0.1 & 0.9 \\

 & 192 & 0.05 & 0.95 & 0.2 & 0.8 & 0.2 & 0.8 & 0.2 & 0.8 & 0.1 & 0.9 \\
 
 & 336 & 0.05 & 0.95 & 0.1 & 0.9 & 0.4 & 0.6 & 0.1 & 0.9 & 0.1 & 0.9 \\
 
 & 720 & 0.05 & 0.95 & 0.1 & 0.9 & 0.4 & 0.6 & 0.1 & 0.9 & 0.1 & 0.9 \\
 \midrule
 
\multirow{4}{*}{\rotatebox[origin=c]{90}{\textbf{ETTm2}}} & 96 & 0.05 & 0.95 & 0.1 & 0.9 & 0.1 & 0.9 & 0.3 & 0.7 & 0.1 & 0.9 \\

 & 192 & 0.05 & 0.95 & 0.1 & 0.9 & 0.1 & 0.9 & 0.2 & 0.8 & 0.01 & 0.99 \\
 
 & 336 & 0.01 & 0.99 & 0.8 & 0.2 & 0.1 & 0.9 & 0.1 & 0.9 & 0.1 & 0.9 \\
 
 & 720 & 0.01 & 0.99 & 0.4 & 0.6 & 0.1 & 0.9 & 0.1 & 0.9 & 0.1 & 0.9 \\
 \midrule
 
\multirow{4}{*}{\rotatebox[origin=c]{90}{\textbf{Exchange}}} & 96 & 0.01 & 0.99 & 0.1 & 0.9 & 0.01 & 0.99 & 0.01 & 0.99 & 0.01 & 0.99 \\

 & 192 & 0.01 & 0.99 & 0.1 & 0.9 & 0.01 & 0.99 & 0.01 & 0.99 & 0.01 & 0.99 \\
 
 & 336 & 0.05 & 0.95 & 0.1 & 0.9 & 0.01 & 0.99 & 0.01 & 0.99 & 0.01 & 0.99 \\
 
 & 720 & 0.05 & 0.95 & 0.1 & 0.9 & 0.01 & 0.99 & 0.01 & 0.99 & 0.1 & 0.9 \\
 \midrule
 
\multirow{4}{*}{\rotatebox[origin=c]{90}{\textbf{Weather}}} & 96 & 0.2 & 0.8 & 0.2 & 0.8 & 0.01 & 0.99 & 0.2 & 0.8 & 0.01 & 0.99 \\

 & 192 & 0.1 & 0.9 & 0.2 & 0.8 & 0.01 & 0.99 & 0.2 & 0.8 & 0.01 & 0.99 \\
 
 & 336 & 0.1 & 1 & 0.2 & 0.8 & 0.01 & 0.99 & 0.1 & 0.9 & 0.2 & 0.8 \\
 
 & 720 & 0.1 & 1 & 0.2 & 0.8 & 0.1 & 0.9 & 0.05 & 0.95 & 0.1 & 0.9 \\ 
 \midrule
 
\multirow{4}{*}{\rotatebox[origin=c]{90}{\textbf{Electricity}}} & 96 & 0.1 & 0.9 & 0.2 & 0.8 & N/A & N/A & 0.2 & 0.8 & 0.1 & 0.9 \\

 & 192 & 0.2 & 0.8 & 0.2 & 0.8 & N/A & N/A & 0.2 & 0.8 & 0.1 & 0.9 \\
 
 & 336 & 0.3 & 0.7 & 0.4 & 0.6 & N/A & N/A & 0.2 & 0.8 & 0.1 & 0.9 \\
 
 & 720 & 0.4 & 0.6 & 0.2 & 0.8 & N/A & N/A & 0.1 & 0.9 & 0.1 & 0.9 \\ 
 \midrule

 \multirow{4}{*}{\rotatebox[origin=c]{90}{\textbf{Traffic}}} & 96 & 0.2 & 0.8 & 0.4 & 0.6 & N/A & N/A & 0.05 & 0.95 & 0.1 & 0.9 \\

 & 192 & 0.2 & 0.8 & 0.4 & 0.6 & N/A & N/A & 0.05 & 0.95 & 0.1 & 0.9 \\
 
 & 336 & 0.2 & 0.8 & 0.4 & 0.6 & N/A & N/A & 0.1 & 0.9 & 0.1 & 0.9 \\
 
 & 720 & 0.1 & 0.9 & 0.5 & 0.5 & N/A & N/A & 0.05 & 0.95 & 0.1 & 0.9 \\
 \midrule
 
\multirow{4}{*}{\rotatebox[origin=c]{90}{\textbf{Illness}}} & 24 & 0.05 & 0.95 & 0.01 & 0.99 & 0.01 & 0.99 & 0.1 & 0.9 & 0.1 & 0.9 \\

 & 36 & 0.05 & 0.95 & 0.01 & 0.99 & 0.5 & 0.5 & 0.1 & 0.9 & 0.1 & 0.9 \\
 
 & 48 & 0.05 & 0.95 & 0.4 & 0.6 & 0.7 & 0.3 & 0.01 & 0.99 & 0.1 & 0.9 \\
 
 & 60 & 0.05 & 0.95 & 0.4 & 0.6 & 0.4 & 0.6 & 0.1 & 0.9 & 0.1 & 0.9 \\
 \bottomrule
\end{tabular}
\caption{The selection of $\{\lambda_{simsia}, \lambda_{pred}\}$ on diverse models. ``+'' denotes the base models that apply the C3RL.
}
\label{lambda_selection_CI}
\end{table*}

\begin{table}[t]
\centering
\begin{tabular}{l|r|cc|cc}
\toprule
\multicolumn{2}{c|}{Model} & \multicolumn{2}{c|}{Informer+} & \multicolumn{2}{c}{Autoformer+}  \\

\multicolumn{2}{c|}{Parameter} & \multicolumn{1}{l}{$\lambda_{simsia}$} &  {$\lambda_{pred}$} & \multicolumn{1}{l}{$\lambda_{simsia}$} &  {$\lambda_{pred}$} \\

\midrule
\multirow{4}{*}{\rotatebox[origin=c]{90}{\textbf{ETTh1}}} & 24(24) & 0.1 & 0.9 & 0.1 & 0.9  \\

 & 48(48) & 0.6 & 0.4 & 0.2 & 0.8  \\
 
 & 168(168) & 0.6 & 0.4 & 0.5 & 0.5  \\
 
 & 336(336) & 0.6 & 0.4 & 0.2 & 0.8  \\

 & 720(720) & 0.6 & 0.4 & 0.4 & 0.6  \\
 \midrule
 
\multirow{4}{*}{\rotatebox[origin=c]{90}{\textbf{ETTh2}}} & 24(24) & 0.2 & 0.8 & 0.3 & 0.7  \\

 & 48(48) & 0.9 & 0.1 & 0.3 & 0.7  \\
 
 & 168(168) & 0.8 & 0.2 & 0.4 & 0.6  \\
 
 & 336(336) & 0.6 & 0.4 & 0.4 & 0.6  \\

 & 720(720) & 0.9 & 0.1 & 0.9 & 0.1  \\
 \midrule
 
\multirow{4}{*}{\rotatebox[origin=c]{90}{\textbf{ETTm1}}} & 24(24) & 0.4 & 0.6 & 0.2 & 0.8  \\

 & 48(48) & 0.1 & 0.9 & 0.2 & 0.8  \\
 
 & 96(96) & 0.6 & 0.4 & 0.2 & 0.8  \\
 
 & 288(288) & 0.1 & 0.9 & 0.1 & 0.9  \\

 & 672(672) & 0.7 & 0.3 & 0.4 & 0.6  \\
 \midrule
 
\multirow{4}{*}{\rotatebox[origin=c]{90}{\textbf{Weather}}} & 24(48) & 0.3 & 0.7 & 0.1 & 0.9  \\

 & 48(96) & 0.9 & 0.1 & 0.1 & 0.9  \\
 
 & 168(192) & 0.7 & 0.3 & 0.2 & 0.8  \\
 
 & 336(336) & 0.8 & 0.2 & 0.2 & 0.8  \\

 & 720(720) & 0.7 & 0.3 & 0.2 & 0.8  \\

 \bottomrule
\end{tabular}
\caption{The selection of $\{\lambda_{simsia}, \lambda_{pred}\}$ on diverse models. ``+'' denotes the base models that apply the C3RL. The forecasting horizon in parentheses is specific to Autoformer.}
\label{lambda_selection_CM}
\end{table}

\end{document}